\documentclass[references=IOPvnum]{iopjournal}

\usepackage{cite}
\usepackage{amsmath,amssymb,amsfonts}
\usepackage{siunitx}
\usepackage{tikz}
\usepackage{pgfplots}
\usetikzlibrary{patterns}
\usepackage{subcaption}
\usepackage{algorithm}
\usepackage{algpseudocode}
\usepackage{makecell}
\usepackage{tabularx}
\usepackage{graphicx}

\pgfplotsset{compat=1.18} 

\begin{document}

\articletype{Paper}

\title{Hyperdimensional Decoding of Spiking Neural Networks}

\author{Cedrick Kinavuidi$^{1,*}$\orcid{0009-0001-2860-9163}, Luca Peres$^1$\orcid{0000-0001-9748-9073} and Oliver Rhodes$^1$\orcid{0000-0003-1728-2828}}

\affil{$^1$ICNS, University of Manchester, Manchester, UK}

\affil{$^*$Author to whom any correspondence should be addressed.}

\email{cedrick.kinavuidi@manchester.ac.uk, luca.peres-2@manchester.ac.uk, oliver.rhodes@manchester.ac.uk}


\begin{abstract}
This work presents a novel spiking neural network (SNN) decoding method, combining SNNs with Hyperdimensional computing (HDC). The goal is to create a decoding method with high accuracy, high noise robustness, low latency and low energy usage. Compared to analogous architectures decoded with existing approaches, the presented SNN-HDC model attains generally better classification accuracy, lower classification latency and lower estimated energy consumption on multiple test cases from literature. The SNN-HDC achieved estimated energy consumption reductions ranging from $1.24\times$ to $3.67\times$ on the DvsGesture dataset and from $1.38\times$ to $2.27\times$ on the SL-Animals-DVS dataset. The presented decoding method can also efficiently identify unknown classes it has not been trained on. In the DvsGesture dataset the SNN-HDC model can identify 100\% of samples from an unseen/untrained class. Given the numerous benefits shown and discussed in this paper, this decoding method represents a very compelling alternative to both rate and latency decoding.
\end{abstract}

\section{Introduction}
Spiking neural networks (SNNs), are the third generation of neural network models \cite{maass_networks_1997}, conceptualised following artificial neural networks (ANNs), but inspired by spiking neurons in the brain. While SNNs have the potential to replace ANNs, partly due to the prospect for significant energy savings \cite{deng_rethinking_2020}, SNNs are less commonly employed as they tend to achieve lower accuracy than ANNs in several learning tasks \cite{roy_towards_2019}, the energy efficiency of SNNs is only realized on specialised hardware \cite{yamazaki_spiking_2022}, and implementing SNNs is more complex than ANNs as SNNs require knowledge of both neuroscience and machine learning \cite{tavanaei_deep_2019}. Neuromorphic algorithms are a key facet of the maturation and wider spread adoption of SNNs. Even though SNNs are meant to be more biologically plausible than ANNs, several of the methods used by SNNs deviate from known biological mechanisms. This includes training using backpropagation through time (BPTT), or simply ignoring the temporal aspect of activity. This paper focuses on improving concept representations and SNN output decoding by developing SNN specific methodology.

By combining SNNs and Hyperdimensional computing (HDC) a new model can be produced which builds on the hypothesis that the brain likely uses distributed representations. This paper will expand on the little but promising research surrounding the combination of SNNs and HDC, and will focus on SNN decoding and how the output layer of an SNN can be changed and interpreted for differing results. We contribute an SNN capable of directly outputting hypervectors which are decoded using HDC. The effectiveness of this model and decoding technique are evaluated against comparable rate and latency decoded models. Performance is analysed through metrics including relative energy consumption, classification latency and classification accuracy as well as other characteristics exclusive to the SNN-HDC model such as the capability of identifying data samples from classes the SNN-HDC has not been trained on.

The paper is structured into the following sections. Section \ref{background} provides an overview of underlying fundamental concepts. Section \ref{related-works} discusses past research related to this work including research on combining SNNs with HDC. Section \ref{methodology} details how SNNs are combined with HDC to create a new architecture, how this architecture encodes data into hypervectors, how it is trained and how experiments were performed. Section \ref{results} presents the experimental results. Section \ref{discusion-conclusion} summarizes the paper and draws conclusions.

\section{Background} \label{background}
\subsection{Spiking Neural Networks}
SNNs aim to be more biologically plausible than ANNs in an attempt to achieve more of the brains benefits, primarily its energy efficiency. Much like the brain, SNNs propagate information using binary values with an inherent temporal aspect called spikes. These spikes are processed in a sparse event driven manner across the entire network. In contrast ANNs typically propagate numerical values and process all neuron activations at synchronous fixed intervals. This difference is a major factor in why SNNs have the potential for superior energy efficiency compared to ANNs \cite{deng_rethinking_2020}. A notable attribute of the brain is consuming energy proportionally to the number of spikes processed \cite{attwell_energy_2001}. This attribute can be attained by pairing neuromorphic hardware with an SNN \cite{caccavella_low-power_2024}. Neuromorphic hardware is a specialized type of hardware inspired by the dynamics observed in the brain with the goal of capturing the brain's attributes, particularly asynchronous event-driven processing \cite{schuman_survey_2017}. 

While it is possible to train an SNN to use very few spikes \cite{davies_advancing_2021}, the most commonly used method, rate decoding \cite{guo_neural_2021}, does the exact opposite as it relies on large amounts of spiking activity to perform well \cite{schuman_evaluating_2022}. Rate decoding is performed by distinguishing between the number of spikes fired by the output neurons over a given time interval/window. This necessitates enough spiking activity inside the SNN such that the SNN can output enough spikes to make a clear distinction between outputs. While this method typically produces the highest accuracy SNNs, the models developed suffer from high latency \cite{guo_neural_2021}, as calculating rates effectively requires accumulating spikes over time. The resulting spiking activity also generates high energy consumption, sometimes requiring more energy than an equivalent ANN \cite{davidson_comparison_2021}. This issue stems from simply translating methods that work for ANNs by making value intensity proportional to the number of spikes instead of developing methods specific to SNNs taking into consideration their temporal properties. 

Looking at studies performed in neuroscience can help to inspire new neuromorphic methodology. The updates that occur in the brain are primarily based on local information \cite{caporale_spike_2008} allowing multiple neurons to react to the same external stimulus. Approximately 85,000 neurons die in the healthy adult brain every day \cite{barrett_optimal_2015}, however it is not the case that people are constantly forgetting various concepts. Given these facts, one can logically conclude that the brain likely represent concepts in a distributed fashion across a large number of neurons. These distributed representations likely exist in a way that allows the brain to achieve low latency \cite{jain_comparative_2015}, low energy usage \cite{furber_build_2012}, and high robustness \cite{barrett_optimal_2015}. 

\subsection{Hyperdimensional Computing} \label{hyperdimensional-computing}
Hyperdimensional computing (HDC) is a brain inspired model that ignores the physical structure of the brain and instead focuses on how the brain can represent and compare concepts with large numbers of neurons \cite{kanerva_hyperdimensional_2009}. HDC performs computations on hypervectors, which are high dimensional vectors comprising the output of a group of neurons, mimicking the distributed representations in the brain. The large number of dimensions, as well as their holographic representation, make them very robust to noise. Figure \ref{fig:hypervectors}, shows two binary hypervectors each with a dimensionality of $10$. The alignment of two hypervectors can be measured to compare them, with high alignment indicating conceptual similarity and low alignment indicating conceptual dissimilarity.

\begin{figure}[H]
    \centering
    \begin{equation*}
        \begin{split}
            A &= \begin{bmatrix} 0 & 1 & 1 & 0 & 1 & 0 & 1 & 1 & 0 & 0 \end{bmatrix} \\
            B &= \begin{bmatrix} 1 & 1 & 0 & 0 & 1 & 0 & 0 & 0 & 0 & 1 \end{bmatrix}
        \end{split}
    \end{equation*}
    \caption{Two random binary hypervectors with $10$ dimensions.}
    \label{fig:hypervectors}
\end{figure}

Cosine similarity, seen in Equation \ref{eq:cosine-similarity}, is a general method for comparing any two hypervectors. Cosine similarity values are within the bounds of $-1 \leq x \leq 1$, with $-1$ representing complete dissimilarity, $0$ representing orthogonality and $1$ representing complete overlap. The cosine similarity of the binary hypervectors $A$ and $B$ from Figure \ref{fig:hypervectors} is $0.447$. Calculating cosine similarity involves performing a dot product which means that dimensions containing zeros do not contribute to the similarity. Calculating more meaningful similarity values requires changing the zeros to $-1$. With this in mind the cosine similarity between hypervectors $A$ and $B$ is $0$.

\begin{equation}
    cos(\theta) = \frac{A \cdot B}{\|A\| \|B\|}
    \label{eq:cosine-similarity}
\end{equation} 

Binary hypervectors can also be compared through a more efficient process called Hamming distance which simply counts the number of mismatches between each dimension (Equation \ref{eq:hamming-distance}). Normalized Hamming distance can be calculated by dividing the Hamming distance by the number of dimensions (Equation \ref{eq:normalised_hamming_distance}). Normalized Hamming distance values are within the bounds of $0 \leq x \leq 1$, with $0$ representing complete overlap, $0.5$ representing orthogonality and $1$ representing complete dissimilarity. Equation \ref{eq:hamming-distance} shows how the Hamming distance is calculated between binary hypervectors $A$ and $B$, which for example from Figure \ref{fig:hypervectors} is $5$. As hypervectors $A$ and $B$ have a dimensionality of $10$, the normalized Hamming distance is $0.5$. Hamming distance can be considered more suitable for neuromorphic hardware where a goal is to minimise energy usage.

\begin{equation}
    HD = \sum_{i=1}^{N} A_i \oplus B_i
    \label{eq:hamming-distance}
\end{equation} 

\begin{equation}
    HD_{\text{norm}} = \frac{1}{N} \sum_{i=1}^{N} A_i \oplus B_i
    \label{eq:normalised_hamming_distance}
\end{equation}

\subsection{SNN Decoding Methods} \label{snn-decoding-methods}
SNN decoding is the process of interpreting the outputs of an SNN to produce insights. As SNNs output spikes over time, both the number and timing of spikes can be used for decoding. Neural networks used for classification trained with supervised learning typically target representations that are one-hot encoded. This involves the output layer having one output neuron for every possible class. The activity desired in each output neuron is dependent on the decoding method. 

Rate decoding classifies inputs based on the output neuron which fires the most spikes. This method is the most commonly seen in SNN research as it typically leads to the highest accuracies. The downsides of this method are its relatively high energy consumption \cite{davidson_comparison_2021} and its high latency \cite{guo_neural_2021}. Equation \ref{eq:mse_loss_rate_decoding} shows how the mean squared error (MSE) of rate decoding can be calculated given $N$ as the number of samples, $r$ as the output normalised firing rates  and $\hat{r}$ as the target normalised firing rates. It is generally believed that rate decoding is not the dominant coding method in the brain, as it does not address the latency observed in the brain \cite{gerstner_neuronal_2014}. 

\begin{equation}
    MSE = \frac{1}{N} \sum_{i=1}^{N}(r_i - \hat{r}_i)^2
    \label{eq:mse_loss_rate_decoding}
\end{equation}

Latency decoding classifies inputs based on the output neuron which fires a spike first. This method generally results in more energy efficient and lower latency SNNs compared to rate decoding \cite{eshraghian_training_2023}, due to fewer spikes getting processed and outputs being trained to fire earlier. However, as it typically achieves lower accuracies \cite{schuman_evaluating_2022} and is less noise robust \cite{guo_neural_2021} it is not as commonly employed. Equation \ref{eq:mse_loss_latency_decoding} shows how the mean squared error (MSE) of latency decoding can be calculated given $N$ as the number of samples, $l$ as the output normalised latencies of the first spike from each neuron and $\hat{l}$ as the target normalised latencies of the first spike from each neuron. It is generally believed that latency decoding is not the dominant coding method in the brain, as it does not address how information is accumulated over time in the brain to make a classification \cite{auge_survey_2021}.

\begin{equation}
    MSE = \frac{1}{N} \sum_{i=1}^{N}(l_i - \hat{l}_i)^2
    \label{eq:mse_loss_latency_decoding}
\end{equation}

Population decoding applies a decoding method using multiple neurons per classification instead of just one, in an effort to mitigate the issues seen in each decoding method. When population decoding is combined with rate decoding, output latency is reduced because rate can be interpreted across multiple output neurons. However, energy consumption increases due to more neurons needing to fire. When population decoding is combined with latency decoding, noise robustness improves as a single erroneous spike cannot by itself dictate the network output. However, latency typically increases as more spikes are required to produce a confident result.

While these methods have been shown to work on samples of data \cite{she_sequence_2021, liu_first-spike_2023}, none of them have addressed inference with continuous data in an event driven manner without introducing additional caveats such as dynamic network resetting \cite{yin_attentive_2022} or adaptive cut-offs \cite{wu_optimizing_2025}. Rate decoding requires counting spikes, raising the question of when counting should start for a classification. Latency decoding relies on the first spike output, raising the question of what classifies as `first' in a system that is always outputting data. While population decoding does address the implausibility of single neuron representations, it does not answer how an SNN could have distributed representations with high accuracy, high noise robustness, low latency and low energy usage.

\subsection{Representation Expressiveness}\label{sub:expressiveness}

One-hot encoding is the prevalent method used in neural networks for representing multi-class categorical data \cite{klimo_deep_2021}. This method involves interpreting the output layer of a neural network as a vector that has one dimension for every classification. Classifications are typically made by simply seeing which dimension has the largest value. In terms of SNNs the largest value could either be the highest number of spikes fired or the earliest spike. As one-hot encoding can only express one class per dimension its expressiveness scales linearly. 

The expressiveness of binary hypervectors does not scale linearly but rather exponentially. Consider a binary hypervector with a $50/50$ chance of every dimension having either binary value, seen in Equation \ref{eq:hypervector_generation}. For a hypervector of size $D$ we will define the expressiveness as the number of hypervectors $N$ that can be generated such that every pair of hypervectors are pseudo-orthogonal. Orthogonal hypervectors have a normalised Hamming distance of $0.5$. Here we will consider pseudo-orthogonal hypervectors to be within the normalised Hamming distance bounds of $0.5 \pm 0.05$. The method to work this out statistically is shown below.

\begin{equation}
    \begin{gathered}
        \mathbf{H} = [H_1, H_2, \ldots, H_D] \\
        \text{for } i = 1, \ldots, D \\
        P(H_i = 1) = 0.5 \\ 
        P(H_i = 0) = 0.5
    \end{gathered}
    \label{eq:hypervector_generation}
\end{equation}

A single dimension in a binary hypervector is a Bernoulli distribution as there are only two possibilities. Both possibilities have $p = 0.5$ chance of occurring. The variance of a single dimension can be calculated as seen in Equation \ref{eq:variance_and_standard_deviation_binary_hypervector}. The variance of a single dimension can then be used to calculate the variance of an entire binary hypervector of arbitrary length $D$. This is then used to calculate the standard deviation of a binary hypervector of length $D$.

\begin{equation}
    \begin{gathered}
        Var(X) = p(1-p) \\
        Var(D_i) = 0.5(1-0.5) \\
        Var(D_i) = 0.25 \\
        Var(D) = D/4 \\ \\
        SD = \sqrt{Var(X)} \\
        SD = \frac{\sqrt{D}}{2}
    \end{gathered}
    \label{eq:variance_and_standard_deviation_binary_hypervector}
\end{equation}

The pseudo-orthogonal bounds defined earlier are Hamming distances within $0.45D$ and $0.55D$. These values, along with a mean of $0.5D$ and the standard deviation (Equation \ref{eq:variance_and_standard_deviation_binary_hypervector}) can be used to calculate the Z scores of the pseudo-orthogonality bounds as shown in Equation \ref{eq:zscore_binary_hypervector}. 

\begin{equation}
    \begin{gathered}
        Z_{\text{score}} = \frac{\text{Raw Score} - \text{Mean}}{\text{Standard Deviation}} \\ \\
        Z_{\text{lower}} = \frac{0.45D - 0.5D}{\sqrt{D} / 2} \\
        Z_{\text{lower}} = -0.1 \sqrt{D} \\ \\ 
        Z_{\text{upper}} = \frac{0.55D - 0.5D}{\sqrt{D} / 2} \\
        Z_{\text{upper}} = 0.1 \sqrt{D}
    \end{gathered}
    \label{eq:zscore_binary_hypervector}
\end{equation}

The hamming distance of any two randomly generated hypervectors follows a normal distribution \cite{kleyko_survey_2022}. The cumulative distribution function can be used to calculate the probability that two randomly generated hypervectors fall within the specified pseudo-orthogonality bounds, seen in Equation \ref{eq:number_of_hypervectors_quadratic}. The probability of a pair of hypervectors being pseudo-orthogonal $P$ is used to calculate the probability of a pair of hypervectors not being pseudo-orthogonal and is expressed as $(1-P)$. The total number of pairs of $N$ hypervectors can be expressed as $\frac{N(N-1)}{2}$. These two expressions can be used together to calculate the number of binary hypervectors that can be generated randomly such that only 1 of the pairs is outside of the previously stated hamming distance bounds. The expression can be rearranged to a quadratic equation solved for the value of $N$. This value is the limit of classes that a hypervector of size $D$ can represent. 

\begin{equation}
    \begin{gathered}
        P = CDF(Z_{\text{upper}}) - CDF(Z_{\text{lower}}) \\
        \frac{N(N-1)}{2} (1-P) = 1 \\
        (1-P)N^2 - (1-P)N - 2 = 0
    \end{gathered}
    \label{eq:number_of_hypervectors_quadratic}
\end{equation}

\begin{figure}[H]
    \centering
    \begin{tikzpicture}
    \begin{axis}[
        width = 0.6\textwidth,
        xlabel={Number of Dimensions},
        ylabel={Number of Classes},
        scaled y ticks = false,
        ymin=1, 
        xmin=1,
        grid=both, 
        legend pos=north west,
    ]
    
    \addplot[
        color=blue,
        line width=1pt,
        smooth
    ]
    coordinates {
        (0, 2)
        (100, 3)
        (200, 4)
        (300, 5)
        (400, 7)
        (500, 9)
        (600, 12)
        (700, 16)
        (800, 21)
        (900, 27)
        (1000, 36)
        (1100, 47)
        (1200, 61)
        (1300, 80)
        (1400, 105)
        (1500, 136)
        (1600, 178)
        (1700, 231)
        (1800, 301)
        (1900, 391)
        (2000, 508)
        (2100, 660)
        (2200, 856)
        (2300, 1111)
        (2400, 1441)
        (2500, 1868)
        (2600, 2420)
        (2700, 3135)
        (2800, 4060)
        (2900, 5257)
        (3000, 6804)
    };
    \addlegendentry{Binary Hypervector}
    
    \addplot[
        color=red,
        line width=1pt,
        dashed, 
    ]
    coordinates {
        (1, 1)
        (3000, 3000)
    };
    \addlegendentry{One-Hot Encoding}
    
    \end{axis}
    \end{tikzpicture}
    \caption{Number of classes that Binary Hypervectors and One-Hot Encoding can represent given a number of dimensions.}
    \label{fig:hypervectors_and_one_hot_encoding}
\end{figure}
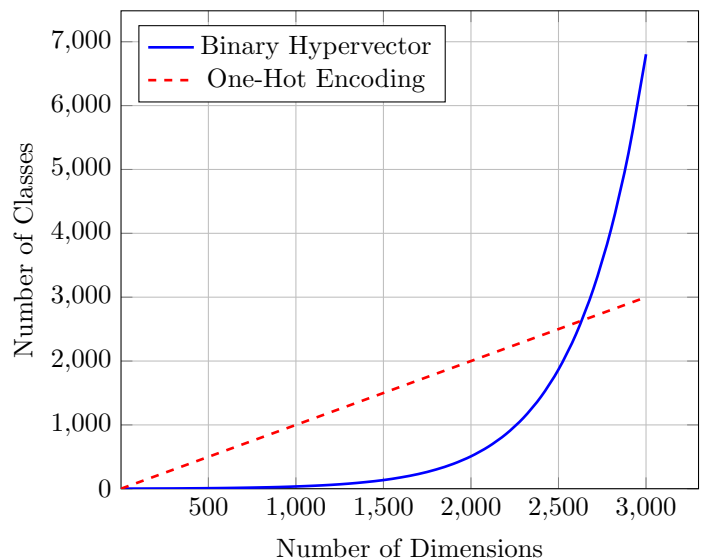

Figure \ref{fig:hypervectors_and_one_hot_encoding} plots the expressiveness of both one-hot encoding and binary hypervectors. It can be seen that for low numbers of dimensions the expressiveness of binary hypervectors is lower than that of one-hot encoding. At a boundary of $D=2633$ the expressiveness of both is approximately the same. Beyond this value the expressiveness of hypervectors increases rapidly and quickly provides more expressiveness than could practically be used. For instance, DeepSeek-V3 (671B) has a vocabulary size of $129,280$ \cite{deepseek-ai_deepseek-v3_2024}. This would require the same number of dimensions when represented with one-hot encoding. A binary hypervector with the same number of dimensions could represent approximately \num{2.2e141} classes. For a binary hypervector to be able to represent $129,280$ classes it would only need $4148$ dimensions. 

\subsection{Neuromorphic Vision}
Neuromorphic vision data is used instead of conventional frame based data as its binary events are a good match to SNNs \cite{deng_rethinking_2020}. Real world neuromorphic vision data is captured using a neuromorphic sensor, such as a dynamic vision sensor (DVS) \cite{lenero-bardallo_36_2011} which mimics the neurobiological structures and functionalities of the retina \cite{liao_neuromorphic_2021}. While a conventional frame-based vision sensor outputs frames (images) which contain data for every pixel, a DVS only outputs events. An event is a threshold change in luminescence that happens at a specific time to a specific pixel. An event stream is series of events over time. In practical terms, a conventional sensor `sees' everything, while a DVS only `sees' movement. Neuromorphic sensors typically offer higher energy efficiency and higher temporal resolution compared to conventional frame-based sensors \cite{liao_neuromorphic_2021}. The DAVIS event based camera \cite{brandli_240_2014} has a resolution of $240\times180$, a latency of $3\,\mathrm{\mu s}$ and a power consumption of $5$ mW to $14$ mW. A comparable conventional high speed frame based camera \cite{chalich_development_2020} has a resolution of $256\times256$, a latency of $429\,\mathrm{\mu s}$ and a power consumption of $4$ W.

\subsection{Temporal Resolution of Neuromorphic Data} \label{resolution_of_data}
SNN models in recent works temporally compress multiple seconds of neuromorphic data into a very small number of timesteps such as $16$ \cite{abad_sneaky_2024}, $20$ \cite{apolinario_s-tllr_2024, sun_synapse-threshold_2023} or $25$ \cite{venkatesh_squat_2024}. This is typically done to speed up training time. While this method is commonly used and the SNNs that use it achieve high accuracies, there are multiple shortcomings with this approach. Firstly, the input frames can cover differing timescales. Using the neuromorphic vision dataset DvsGesture \cite{amir_low_2017} as an example, the shortest sample in the dataset is 1.7 s while the longest is 18.5 s. If both of these samples are compressed into 25 frames, then one frame in the longer sample contains more time than a frame in the shorter sample. Given the knowledge that SNNs make use of temporal information \cite{eshraghian_training_2023}, raises the question of the appropriateness of inputting differing timescales as well as what timescale one should use for inference of a data sample with an unknown length. Secondly, as data is being compressed temporally, there is a risk that temporal information is being diminished. Thirdly, when it comes to real world deployment, compressing frames has a significantly negative impact on the latency of the model. A real world sample that takes ten seconds to capture cannot be processed completely in any less than ten seconds. Consider a model that is trained on data samples which are 10 s long compressed into 25 equal frames. Should this model be deployed in the real world it could only produce an output once every 400 ms. Compare this to a model that is trained with consistent 1 millisecond frames. This model would be able to produce an output once every millisecond. Deployed in the real world both of these models would take the same amount of time to fully process each data sample, however, since the latter model has a higher temporal resolution, it has the potential for lower latency. It is ideal to move towards solutions that are tailored towards real world performance as opposed to just improving dataset accuracy.

\section{State-of-the-art} \label{related-works}
While there is relatively limited research on the topic of combining SNNs and HDC, the little research that does exist is promising.

SpikeHD \cite{zou_memory-inspired_2022} developed a convolutional SNN which was combined with HDC. The authors began with a convolutional SNN which was trained using Deep Continuous Local Learning (DECOLLE) \cite{kaiser_synaptic_2020} using smooth L1 loss. The SNN was trained on neuromorphic data including the DvsGesture dataset \cite{amir_low_2017} and MNIST dataset \cite{deng_mnist_2012} (Poisson generated spikes). After this training took place the last layer of the SNN was removed, turning it into a feature extractor. After data samples were passed through this feature extractor, the outputs were turned into hypervectors using projection matrices. These hypervectors were then used to train an HDC model. After the HDC model was trained the authors compared the performance of the full SNN and the feature extractor combined with the HDC. Their results showed that the combined model was able to achieve 5.7\% and 3.2\% higher classification accuracy than the SNN by itself on MNIST \cite{deng_mnist_2012} and DvsGesture \cite{amir_low_2017}, respectively. The authors believe that this two stage information processing is what caused the performance improvement.

HyperSpike \cite{morris_hyperspike_2022}, showcased a single layer untrained SNN that when combined with HDC was able to achieve results comparable to surrogate gradient trained SNNs. After a data sample was passed through the untrained SNN, the outputs were turned into hypervectors using projection matrices. These hypervectors were used to train the HDC model, this combined model was then compared to SNNs trained on a range of neuromorphic vision datasets. The combined model achieved 0.2\% higher on the N-MNIST dataset \cite{orchard_converting_2015}, 6.6\% lower on the DvsGesture dataset \cite{amir_low_2017} and 1.5\% lower on the ASL-DVS dataset \cite{bi_graph-based_2019} compared to the surrogate gradient trained SNNs. This shows that HDC is effective at extracting information from an SNN, even though the SNN has not undergone any training.

The research described above shows promising results but there are potential improvements to be made. SpikeHD did not initially train its SNN directly on hypervector representations of each class, meaning the inner dynamics of the SNN were optimised to target a different representation of the classes. The resulting combined model was later updated by what can essentially be described as transfer learning. Not only does this increase training time and cost, transfer learning can result in worse performance than models that are trained from scratch \cite{galvez_yolo-based_2019}. It is possible that a combined model trained to target hypervectors initially would have better performance. While HyperSpike did not train the SNN and achieved good performance, training the SNN portion would result in better performance \cite{li_computational_2018} due to the feature extractor being more effective.

Both SpikeHD and HyperSpike require matrix multiplications to encode the data output by the SNN into hypervectors. While matrix multiplication has been optimised on GPUs, avoiding the reintroduction of matrix multiplication helps to maintain the benefits of neuromorphic hardware. One way to make encoding more energy efficient, which could also be considered more neuromorphic, would be to remove any multiplication and have it based solely on addition as multiplication requires considerably more energy than addition \cite{horowitz_11_2014}. A $45$ nm processor requires $0.1$ pJ for a 32 bit addition and $3.1$ pJ for a 32 bit multiplication \cite{horowitz_11_2014}.

Both papers pass the entire data sample through the SNN before encoding the SNN output into a hypervector, negatively impacting the latency at which predictions can be made. No predictions can be made before the full length of a sample has been observed, contrary to both rate and latency decoding which can produce confident outputs in the middle of observing a sample. Using a hypervector encoding method that can be applied in an event driven manner would reduce latency and enable continuous predictions.

\section{Methodology} \label{methodology}
\begin{figure}[H]
    \centering
    \begin{subfigure}{0.7\textwidth}
        \includegraphics[width=\textwidth]{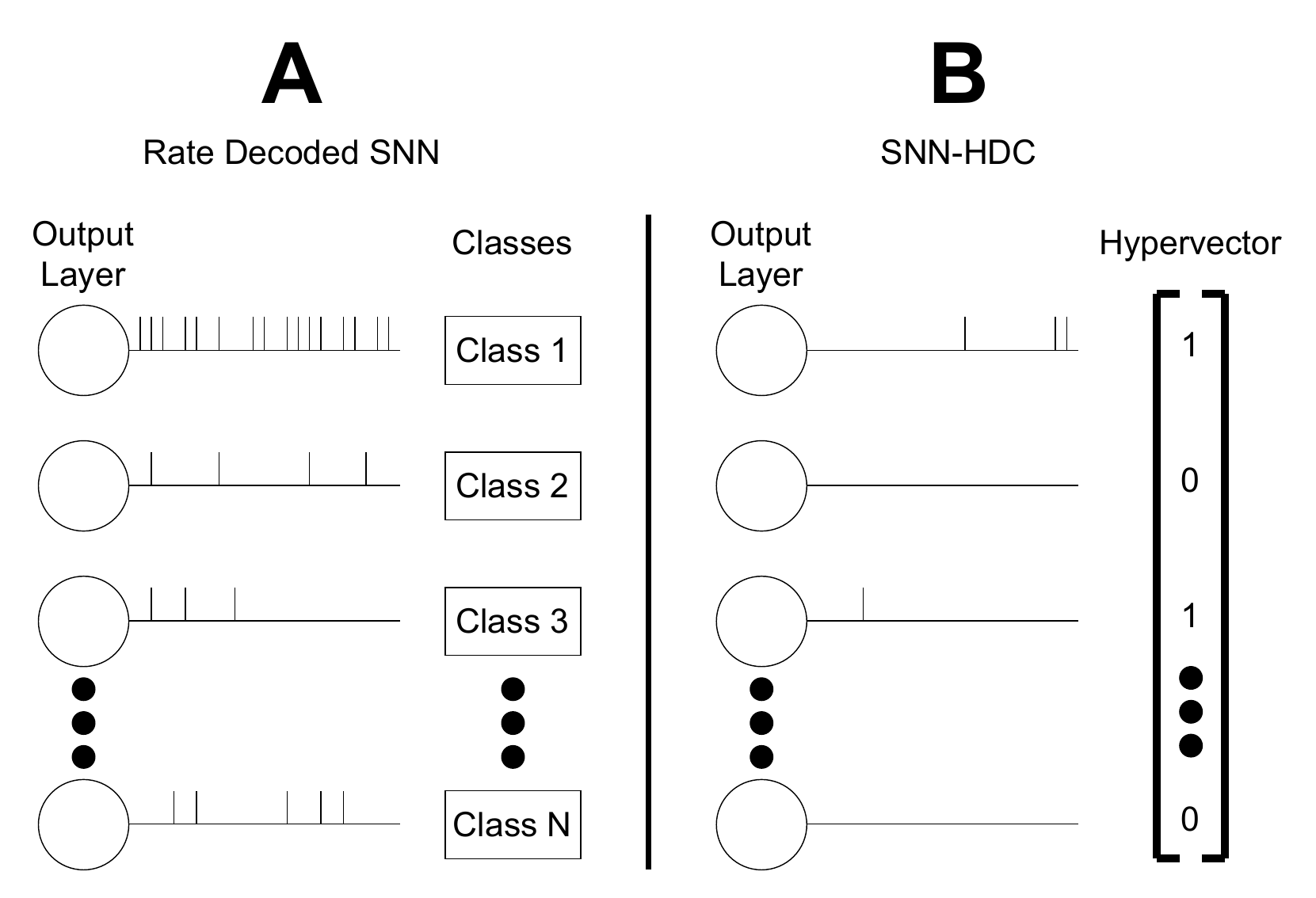}
        \phantomsubcaption
        \label{fig:snn-comparisonA}
    \end{subfigure}
    \begin{subfigure}{0pt}
        \phantomsubcaption
        \label{fig:snn-comparisonB}
    \end{subfigure}
    \caption{A rate decoded SNN (A) and the SNN-HDC model (B). The rate decoded SNN has one output neuron per classification. Inputs are classified based on highest spike count. Here the input is classified as Class 1. The SNN-HDC has an arbitrary amount of output neurons used to build a hypervector over time. Every dimension initialises with a value of $0$. This value is changed to $1$ if that dimension observes the presence of spikes.}
    \label{fig:snn-comparison}
\end{figure}

The difference between a standard rate decoded SNN and the proposed model in this paper can be seen in Figure \ref{fig:snn-comparison}. The rate decoded SNN (Figure \ref{fig:snn-comparisonA}) has one output neuron per class (one-hot encoding). The model proposed here, termed SNN-HDC (Figure \ref{fig:snn-comparisonB}), has an experimentally determined number of output neurons. The proposed encoding process uses the presence and absence of output spikes as binary dimensional values. Each of the output neurons corresponds to its own dimension in the hypervector and each dimension is initialized to zero. A neuromorphic data sample captured with a dynamic vision sensor (DVS) \cite{lenero-bardallo_36_2011} is passed through the SNN. Simultaneously with the data input, the presence of spikes from an output neuron causes the corresponding dimension to be flipped from zero to one. This hypervector is then compared with known hypervectors using Hamming distance (Equation \ref{eq:hamming-distance}) to produce classifications. 

The class hypervectors are generated before training using Equation \ref{eq:hypervector_generation}. The class hypervectors are all binary and are generated with each dimension having a $50\%$ chance to be either $0$ or $1$. While this generation is random, due to the high number of dimensions that hypervectors possess, it results in hypervectors that are all pseudo-orthogonal to each other \cite{kanerva_hyperdimensional_2009}. Binary hypervectors were chosen over non-binary hypervectors as they allow for more energy efficient comparisons as discussed in Section \ref{hyperdimensional-computing}. This method presumes that output neurons react to very specific temporal features and that the frequency and latency of their spiking patterns is not as important as their presence.

\subsection{Neuromorphic Data}\label{sub:datasets}
This paper will utilise multiple neuromorphic datasets to explore how the method responds to different types of data. All of the data in this work will utilise one millisecond long frames. This ensures temporal consistency of inputs, reduces the risk of diminishing temporal information and increases the output temporal resolution of each model. 

\subsubsection{DvsGesture}
The DvsGesture dataset \cite{amir_low_2017} contains eleven classes of various hand gestures, with a total of $1341$ data samples. The Tonic \cite{lenz_tonic_2021} library was used to load the data and provided its train/test split of $1077/264$, with $24$ samples of each class composing the test set. The data was downsampled to $32 \times 32$ and only the first $1500$ ms of each data sample are used to account for the wide discrepancy between temporal lengths of samples and to speed up training time. 

\subsubsection{SL-Animals-DVS}
The SL-Animals-DVS dataset \cite{vasudevan_sl-animals-dvs_2022} contains nineteen classes of various sign language words of animals being performed. A total of $59$ people performed each of the $19$ signs under four different lighting conditions. The dataset paper \cite{vasudevan_sl-animals-dvs_2022} contains results for models trained on all four lighting conditions as well as separate models trained on a subset of three lighting conditions. This was done as one of the lighting conditions introduces significant noise to the data samples making classification harder \cite{vasudevan_sl-animals-dvs_2022}. All results presented in this paper use all four lighting conditions to evaluate performance on all real world  conditions which includes noisy samples. As there is no default train/test split, a strategy of K-fold cross validation with a leave-signers-out approach was used, duplicating the methodology used by the dataset paper \cite{vasudevan_sl-animals-dvs_2022}. The data was downsampled to $32 \times 32$ and only the first $1500$ ms of each sample is used as the dataset paper \cite{vasudevan_sl-animals-dvs_2022} states this time frame allows every sign to be performed and avoids overlaps. 

\subsection{SNN Architecture}
The models defined in this work use a relatively small number of layers/parameters compared to SNN models that currently achieve the highest accuracies on both the DvsGesture dataset \cite{she_sequence_2021, apolinario_s-tllr_2024, sun_synapse-threshold_2023} and the SL-Animals-DVS dataset \cite{sun_eventrpg_2024, sabater_event_2022}. The goal of this paper is not to develop large models to attain state-of-the-art accuracy but instead to compare the effectiveness of different SNN decoding techniques. Using a small model that has already been verified as being effective allows for faster experimentation while facilitating thorough analysis of SNN decoding techniques.

Different architectures will be used for each of the neuromorphic datasets to show that the proposed decoding method consistently works regardless of the architecture used. All the models used in this paper are convolutional SNNs with leaky-integrate-and-fire (LIF) neurons. 

The SNN-HDC method proposed in this work requires not only algorithmic changes for loss calculation and inference but also architectural changes, namely increasing the number of the neurons in the output layer, relative to a standard rate/latency decoded SNN. Due to these changes in architecture, the SNN-HDC is compared with multiple analogous SNNs for a fair analysis. Three types of model are used for each dataset: the SNN-HDC; a model of the same depth (SNN-1); and a model that is one layer deeper than the SNN-HDC (SNN-2). Tables \ref{tab:gestures-architectures} and \ref{tab:animals-architectures} show the architectural differences between the three model types and are further detailed below.

The difference between the SNN-HDC and the same depth model is just in the output layer. The SNN-HDC model will have an experimentally determined number of output neurons whilst the same-depth model is one-hot encoded. The deeper model copies the architecture of the SNN-HDC, and appends a one-hot encoded output layer. Comparing the SNN-HDC to an architecture of the same depth will show what happens when an SNN output layer is replaced by a hyperdimensional layer. Comparing the SNN-HDC to a deeper architecture will offer insights around whether the performance differences are due to increasing the number of neurons in the output layer. The same depth (SNN-1) and deeper (SNN-2) architectures are trained using both rate and latency decoding resulting in four SNNs which will be referred to as Rate-1, Rate-2, Latency-1 and Latency-2.

\begin{table}[H]
    \centering
    \caption{Architectures used for the DvsGesture dataset \cite{amir_low_2017}. $D$ refers to a fully connected layer with $D$ number of LIF neurons. SNN-1 and SNN-2 are trained with rate and latency decoding giving 5 networks in total: SNN-HDC, Rate-1, Rate-2, Latency-1 and Latency-2.}
    \begin{tabular}{|| c | c | c ||}
        \hline 
        SNN-HDC & SNN-1 (Same Depth) & SNN-2 (Deeper)\\ [0.5ex]
        \hline\hline
        \makecell{16Conv5 \\ BatchNorm \\ Maxpool \\ Dropout \\ 32Conv5 \\ BatchNorm \\ Maxpool \\ Dropout \\ $D$ \\ \ } & 
        \makecell{16Conv5 \\ BatchNorm \\ Maxpool\\ Dropout \\ 32Conv5 \\ BatchNorm \\ Maxpool \\ Dropout \\ $11$ \\ \ } & 
        \makecell{16Conv5 \\ BatchNorm \\ Maxpool \\ Dropout \\ 32Conv5 \\ BatchNorm \\ Maxpool \\ Dropout \\ $D$ \\ $11$} \\ 
        \hline
    \end{tabular}
    \label{tab:gestures-architectures}
\end{table}

The DvsGesture SNN-HDC uses an architecture of [$16c5$ - $bn$ - $2p$ - $0.2d$ - $32c5$ - $bn$ - $2p$ - $0.2d$ - $D$] where $c$ denotes a convolutional layer, $bn$ denotes batch normalisation, $p$ denotes a maxpool operation, $d$ denotes a dropout layer and $D$ denotes a fully connected output layer with $D$ number of neurons. This architecture is based on recently presented work on the same dataset \cite{venkatesh_squat_2024}. The Rate-1 and Latency-1 networks  replace $D$ with $11$ neurons. The Rate-2 and Latency-2 networks append $11$ neurons after $D$. The $11$ neurons form the one-hot encoded output layer. The model architectures are summarised in Table \ref{tab:gestures-architectures}. The performance of the baseline model, this work and other published models can be seen in Table \ref{tab:gestures_performance_from_other works}. 

The SL-Animals-DVS SNN-HDC uses an architecture of [$8c5$ - $2$p - $16c5$ - $2p$ - $32c5$ - $2p$ - $25fc$ - $D$] where $c$ denotes a convolutional layer, $p$ denotes a maxpool operation, $fc$ denotes a fully connected layer and $D$ denotes a fully connected output layer with $D$ number of neurons. This architecture is based on the architecture presented in the original dataset paper \cite{vasudevan_sl-animals-dvs_2022} albeit with smaller and fewer convolutional kernels due to available resources. The Rate-1 and Latency-1 networks replace $D$ with $19$ neurons. The Rate-2 and Latency-2 networks append $19$ neurons after $D$. The $19$ neurons form the one-hot encoded output layer. The model architectures are summarised in Table \ref{tab:animals-architectures}. The performance of the baseline model, this work and other published models can be seen in Table \ref{tab:animals_performance_from_other works}.

\begin{table}[H]
    \centering
    \caption{Architectures used for the  SL-Animals-DVS dataset \cite{vasudevan_sl-animals-dvs_2022}. $D$ refers to a fully connected layer with $D$ number of LIF neurons. SNN-1 and SNN-2 are trained with rate and latency decoding giving 5 networks in total: SNN-HDC, Rate-1, Rate-2, Latency-1 and Latency-2.}
    \begin{tabular}{|| c | c | c ||}
        \hline 
        SNN-HDC & SNN-1 (Same Depth) & SNN-2 (Deeper)\\ [0.5ex]
        \hline\hline
        \makecell{8Conv5 \\ Maxpool \\ 16Conv5 \\ Maxpool \\ 32Conv5 \\ Maxpool \\ $25$ \\ $D$ \\ \ } & 
        \makecell{8Conv5 \\ Maxpool \\ 16Conv5 \\ Maxpool \\ 32Conv5 \\ Maxpool \\ $25$ \\ $19$ \\ \ } & 
        \makecell{8Conv5 \\ Maxpool \\ 16Conv5 \\ Maxpool \\ 32Conv5 \\ Maxpool \\ $25$ \\ $D$ \\ $19$} \\ 
        \hline
    \end{tabular}
    \label{tab:animals-architectures}
\end{table}

The LIF neurons used by all models in this work are described in Equations \ref{eq:lif_intergrate_and_leak} and \ref{eq:neuron_fire_reset}. Equation \ref{eq:lif_intergrate_and_leak} shows the synaptic integration and leakage of a LIF neuron. The membrane voltage at the current timestep $t$, is defined as $M_i[t]$, the decay rate is defined as $\beta$, the synaptic weight matrix is defined as $W_{ij}$, bias values defined as $I_i$ and the presence or absence of input spikes at the current timestep is defined as $X_j[t]$. The value of $\beta$ is determined experimentally for each model. Equation \ref{eq:neuron_fire_reset} shows that a LIF neuron fires a spike and resets its membrane to $0$ if the membrane reaches the threshold value of $1$. The snnTorch library \cite{eshraghian_training_2023} is used for training. The LIF neurons make use of the library's reset delay functionality which makes an LIF neuron propagate spikes one timestep after the membrane threshold has been reached. This is used to prevent spikes passing through multiple layers of the SNNs in a single timestep. This is useful when designing implementations for neuromorphic hardware, which typically require at least one timestep to route spikes between neurons \cite{furber_overview_2013}.

\begin{equation}
    M_i[t] = \beta M_i[t-1] + W_{ij}X_j[t] + I_i
    \label{eq:lif_intergrate_and_leak}
\end{equation}

\begin{equation}
    M_i[t], X_i[t] =
    \begin{cases}
    (0, 1), & \text{if } M_i[t] \geq 1, \\
    (M_i[t], 0), & \text{otherwise.}
    \end{cases}
    \label{eq:neuron_fire_reset}
\end{equation}

\subsection{Model Training}
The snnTorch library \cite{eshraghian_training_2023} is used to train the SNN models, with all models trained using surrogate gradients and backpropagation through time. All models are trained on the datasets described in section \ref{sub:datasets}, with a batch size of $32$ and the default Adam parameters \cite{kingma_adam_2017}. All models are trained across three different runs using three different PyTorch \cite{ansel_pytorch_2024} seeds affecting the weight initialisation of all models and the generated class hypervectors for the SNN-HDC models. The difference between the models (besides architecture) is the way that the model output is used to determine the loss. 

The rate decoded models are trained to target an 80\% firing rate for the correct output neuron and a 20\% firing rate for all incorrect output neurons, which are commonly used targets in snnTorch implementations \cite{eshraghian_training_2023}. The firing rates here indicate the percentage of timesteps that an output neuron should fire. The normalised firing rate is used to calculate the mean squared error (MSE) loss (Equation \ref{eq:mse_loss_rate_decoding}). 

The latency decoded models are trained to fire a spike in the correct output neuron at the earliest possible timestep and for the incorrect output neurons at the last possible timestep. The normalised timing relative to the duration of the entire input is used to calculate the MSE loss (Equation \ref{eq:mse_loss_latency_decoding}).

The loss for the SNN-HDC is calculated using MSE loss in Equation \ref{eq:mse_loss_snnhdc} given $N$ as the number of samples, $H$ as the output hypervector and $C$ as the target class hypervector. $H$ is the sum of the output spikes with the desired `on' dimensions clamped to $1$ as shown in Algorithm \ref{alg:snn_hdc_loss_calculation}. The proposed hypervector encoding process is based on the presence and absence of spikes. The SNN-HDC model has $D$ output neurons encoding a hypervector with dimensionality of $D$. The $i^{th}$ output neuron corresponds to the $i^{th}$ dimension in the hypervector. Every dimension in the hypervector is initialised with $0$. Should the $i^{th}$ output neuron spike during the input of a data sample, the corresponding $i^{th}$ dimension will be flipped from $0$ to $1$. If the dimensions in the hypervector are thought of as flags corresponding to specific temporal features then a dimension spiking multiple times means that specific temporal feature was observed multiple times. For a dimension where the target is a $1$, loss does not increase for a spike count above one. For a dimension where the target is a $0$, loss increases proportionally to the spike count.  

\begin{equation}
    MSE = \frac{1}{N} \sum_{i=1}^{N}(H_i - C_i)^2
    \label{eq:mse_loss_snnhdc}
\end{equation}

\begin{algorithm}
    \centering
    \begin{algorithmic}
        \State $H \gets \sum (spikes \ output \ by \ each \ neuron)$
        \State $C \gets$ Target class hypervector
        \For{\texttt{$index$ \textbf{in} $H$}}
            \If{$C[index] = 1$}
                \If{$H[index] > 1$}
                    \State $H[index] \gets 1$
                \EndIf
            \EndIf
        \EndFor
        \State $Loss \gets MSE(H, C)$
    \end{algorithmic}
    \caption{Calculating loss of SNN-HDC model.}
    \label{alg:snn_hdc_loss_calculation}
\end{algorithm}

\subsection{Synaptic Operations and Estimated Energy Consumption}\label{sub:estimated-energy}
A common goal of neuromorphic hardware is attempting to consume energy relative to the number of synaptic operations (SOPs) that take place. A SOP corresponds to a source neuron sending a spike event to a target neuron via a unique (non-zero) synapse \cite{merolla_million_2014}. The TrueNorth neuromorphic hardware uses $26$ pJ per SOP \cite{merolla_million_2014}. The energy values shown here are estimates of the power consumption of each model per sample of data that it processes. Every test set sample for a given dataset is passed through the trained SNNs. The number of SOPs occurring in each SNN is counted and the average number of SOPs per sample is calculated. This SOP value is multiplied by $26$ pJ to produce an energy consumption estimate.

\subsection{Membrane Potential Decay Rate and Dimensionality}
The rate at which the membrane potential decays in a neuron affects the overall activity in the network. As the optimal decay rate may be different for different decoding techniques, every decoding technique will be trained using multiple values and the optimal value for a given architecture will be used for the final analysis. The membrane potential decay rate is defined as $\beta$ in Equation \ref{eq:lif_intergrate_and_leak}. Values of $\beta$ that will be tested are $0.5$, $0.7$, and $0.9$. The values of $0.5$ and $0.9$ were chosen as they are commonly seen $\beta$ values in snnTorch \cite{eshraghian_training_2023} implementations, with $0.7$ chosen as an intermediate value. 

The dimensionality value used by the models is chosen experimentally by analysing the accuracy and synaptic operations of various dimensionality values. The dimensionality values tested range from $11$ (giving the SNN-HDC model the same number of output neurons as the rate and latency models for the DvsGesture dataset \cite{amir_low_2017}) up to a value where the accuracy no longer appears to benefit from higher dimensionality.

The DvsGesture dataset \cite{amir_low_2017} will act as an exploratory stage to determine an optimal combination of membrane potential decay rate and dimensionality. These values will be fixed for the SL-Animals-DVS dataset \cite{vasudevan_sl-animals-dvs_2022}.

\subsection{Measuring Latency}
The latency of the models determines how quickly classifications are made. As three different decoding methods are being used, three different ways of measuring latency must be used. For rate decoding the latency of an output is defined as how long it takes the network to reach a softmax of at least $0.99$ across the sum of the output neuron spikes, as suggested by recent work \cite{li_seenn_2023}. For latency decoding, the latency is defined as the time taken for the network to output its first spike. For hyperdimensional decoding the latency is defined as the time taken for the network to settle on an answer that doesn't change for the remaining duration of the input sample. Latency is calculated by passing all test samples into the trained networks, and reported as average $\pm$ standard deviation, for each dataset.

\subsection{Identifying Unknown Classes}
Unlike rate and latency decoded models, the SNN-HDC is not inherently constrained to the number of classes it can output. A rate or latency decoded SNN with eleven output neurons can only output eleven classes. As an SNN-HDC outputs hypervectors instead of classes, regardless of how many classes it has been trained on, new classes can be identified if the output hypervector does not share similarity with any known class hypervector. This ability to detect unknown classes is explored using the DvsGesture dataset \cite{amir_low_2017}. This dataset contains eleven different classes, ten of which are assigned concepts and one of which is a random concept. After the optimal $\beta$ and dimensionality parameters have been chosen for the eleven class SNN-HDC, the same parameters will be used to train another SNN-HDC model on a subset of the dataset containing only the ten assigned concepts. This model is then evaluated on all eleven concepts, identifying the random concepts based on their dissimilarity to the known class hypervectors. 

The eleven class SNN-HDC models classify inputs purely based on similarity, however, the ten class SNN-HDC model will classify inputs based on a minimum level of similarity. The ten class SNN-HDC will classify an input as a known class if the normalized Hamming Distance is less than $\delta$, otherwise the input will be classified as an unknown class. The values of $\delta$ used for analysis are within the range of $0.1$ to $0.4$, representing high alignment and low alignment. The accuracy for both the whole dataset as well as just the eleventh class is tracked. The performance of the ten class SNN-HDC is also compared to the eleven class SNN-HDC.

\section{Results} \label{results}
The results section presents the outcomes of the experiments outlined in the methodology section. An analysis of the SNN-HDC models is performed in subsection \ref{sub:snnhdc_analysis}, followed by an analysis of the rate and latency decoded models in subsection \ref{sub:rate_latency_analysis}. After this analysis, one model is chosen for each of the five different model types. The model activity, including the number of spikes, estimated energy consumption and classification latency are then compared in subsections \ref{sub:model_activity} and \ref{sub:model_latency}. The ability of the SNN-HDC model to identify classes that it has not been trained on is also tested in subsection \ref{sub:unknown_class}. Detailed exploratory experimentation of the DvsGesture dataset \cite{amir_low_2017} is shown initially to determine the best parameters, with subsequent datasets only showing the key highlights. Finally, the models produced in this research are compared to models from other research that have been trained on the same dataset in subsections \ref{sub:gesture_results} and \ref{sub:animals_results}. 

\subsection{SNN-HDC Analysis} \label{sub:snnhdc_analysis}
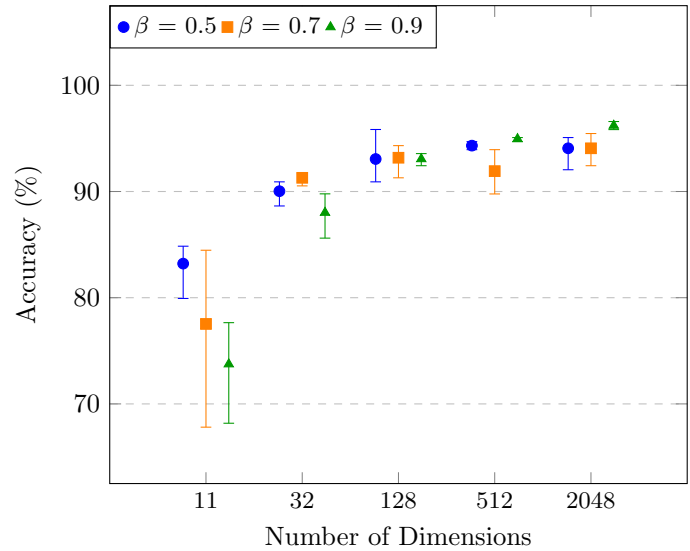
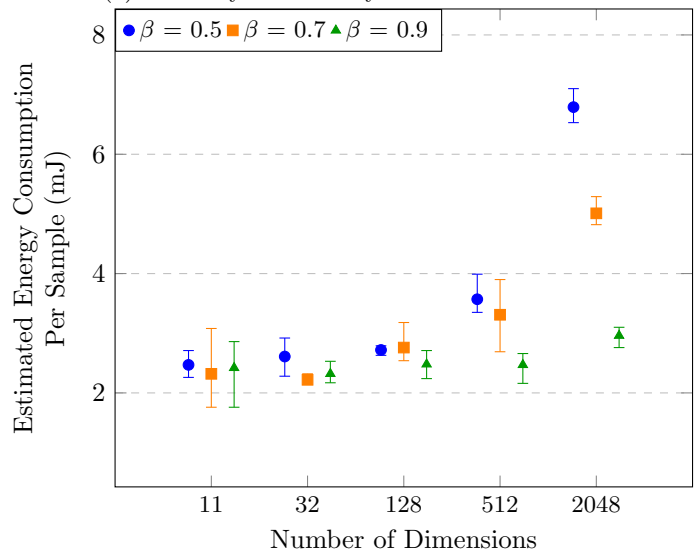
\begin{figure}[H]
  \centering
  \begin{subfigure}{\textwidth}
    \centering
    \begin{tikzpicture}
        \begin{axis}[
            width  = 0.6\textwidth,
            ymin=70, 
            ymax=100, 
            ylabel={Accuracy (\%)}, 
            xlabel={Number of Dimensions},
            xtick={0,1,2,3,4}, 
            xticklabels={11, 32, 128, 512, 2048}, 
            xticklabel style={
                font=\small,
            },
            enlargelimits=0.25, 
            ymajorgrids, 
            grid style=dashed, 
            nodes near coords, 
            nodes near coords style={font=\tiny, black, anchor=south, text opacity=0}, 
            legend style={at={(0,1)}, anchor=north west, legend columns=3, font=\small},
        ]

        \addplot[
            scatter, only marks, mark=*, mark size=2pt, color=blue,
            xshift=-3mm, 
            error bars/.cd, y dir=both, y explicit, error bar style={color=blue}, error mark=-,
        ] coordinates {
            (0, 83.21) += (0, 1.64) -= (0, 3.28)
        };
        \addplot[
            scatter, only marks, mark=square*, mark size=2pt, color=orange,
            xshift=0mm, 
            error bars/.cd, y dir=both, y explicit, error bar style={color=orange}, error mark=-,
        ] coordinates {
            (0, 77.53) += (0, 6.94) -= (0, 9.72)
        };
        \addplot[
            scatter, only marks, mark=triangle*, mark size=2pt, color=green!60!black,
            xshift=3mm, 
            error bars/.cd, y dir=both, y explicit, error bar style={color=green!60!black}, 
            error mark=-,
        ] coordinates {
            (0, 73.74) += (0, 3.91) -= (0, 5.56)
        };

        \addplot[
            scatter, only marks, mark=*, mark size=2pt, color=blue,
            xshift=-3mm, 
            error bars/.cd, y dir=both, y explicit, error bar style={color=blue}, error mark=-,
        ] coordinates {
            (1, 90.03) += (0, 0.88) -= (0, 1.39)
        };
        \addplot[
            scatter, only marks, mark=square*, mark size=2pt, color=orange,
            xshift=0mm, 
            error bars/.cd, y dir=both, y explicit, error bar style={color=orange},  error mark=-,
        ] coordinates {
            (1, 91.29) += (0, 0.38) -= (0, 0.76)
        };
        \addplot[
            scatter, only marks, mark=triangle*, mark size=2pt, color=green!60!black,
            xshift=3mm, 
            error bars/.cd, y dir=both, y explicit, error bar style={color=green!60!black}, 
            error mark=-,
        ] coordinates {
            (1, 88.01) += (0, 1.77) -= (0, 2.4)
        };

        \addplot[
            scatter, only marks, mark=*, mark size=2pt, color=blue,
            xshift=-3mm, 
            error bars/.cd, y dir=both, y explicit, error bar style={color=blue}, error mark=-,
        ] coordinates {
            (2, 93.06) += (0, 2.78) -= (0, 2.15)
        };
        \addplot[
            scatter, only marks, mark=square*, mark size=2pt, color=orange,
            xshift=0mm, 
            error bars/.cd, y dir=both, y explicit, error bar style={color=orange},  error mark=-,
        ] coordinates {
            (2, 93.18) += (0, 1.14) -= (0, 1.89)
        };
        \addplot[
            scatter, only marks, mark=triangle*, mark size=2pt, color=green!60!black,
            xshift=3mm, 
            error bars/.cd, y dir=both, y explicit, error bar style={color=green!60!black}, 
            error mark=-,
        ] coordinates {
            (2, 93.06) += (0, 0.51) -= (0, 0.63)
        };

        \addplot[
            scatter, only marks, mark=*, mark size=2pt, color=blue,
            xshift=-3mm, 
            error bars/.cd, y dir=both, y explicit, error bar style={color=blue}, error mark=-,
        ] coordinates {
            (3, 94.32) += (0, 0.38) -= (0, 0.38)
        };
        \addplot[
            scatter, only marks, mark=square*, mark size=2pt, color=orange,
            xshift=0mm, 
            error bars/.cd, y dir=both, y explicit, error bar style={color=orange},  error mark=-,
        ] coordinates {
            (3, 91.92) += (0, 2.02) -= (0, 2.15)
        };
        \addplot[
            scatter, only marks, mark=triangle*, mark size=2pt, color=green!60!black,
            xshift=3mm, 
            error bars/.cd, y dir=both, y explicit, error bar style={color=green!60!black}, 
            error mark=-,
        ] coordinates {
            (3, 94.95) += (0, 0.13) -= (0, 0.25)
        };

        \addplot[
            scatter, only marks, mark=*, mark size=2pt, color=blue,
            xshift=-3mm, 
            error bars/.cd, y dir=both, y explicit, error bar style={color=blue}, error mark=-,
        ] coordinates {
            (4, 94.07) += (0, 1.01) -= (0, 2.02)
        };
        \addplot[
            scatter, only marks, mark=square*, mark size=2pt, color=orange,
            xshift=0mm, 
            error bars/.cd, y dir=both, y explicit, error bar style={color=orange},  error mark=-,
        ] coordinates {
            (4, 94.07) += (0, 1.39) -= (0, 1.64)
        };
        \addplot[
            scatter, only marks, mark=triangle*, mark size=2pt, color=green!60!black,
            xshift=3mm, 
            error bars/.cd, y dir=both, y explicit, error bar style={color=green!60!black}, 
            error mark=-,
        ] coordinates {
            (4, 96.21) += (0, 0.38) -= (0, 0.38)
        };
    
        \addlegendentry{$\beta$ = 0.5}
        \addlegendentry{$\beta$ = 0.7}
        \addlegendentry{$\beta$ = 0.9}
        
        \end{axis}
        
    \end{tikzpicture}

    \caption{Accuracy obtained by SNN-HDC models.}
    \label{fig:snn-hdc-dimensionality-beta-accuracy}
  \end{subfigure}
  \hfill
  \begin{subfigure}{\textwidth}
    \centering
    \begin{tikzpicture}
        \begin{axis}[
            width  = 0.6\textwidth,
            ylabel={\makecell{Estimated Energy Consumption \\ Per Sample (mJ)}}, 
            scaled y ticks = false,
            xtick={0,1,2,3,4}, 
            xlabel={Number of Dimensions},
            xticklabels={11, 32, 128, 512, 2048}, 
            xticklabel style={
                font=\small,
            },
            enlargelimits=0.25,
            ymajorgrids, 
            grid style=dashed, 
            nodes near coords, 
            nodes near coords style={font=\tiny, black, anchor=south, text opacity=0}, 
            legend style={at={(0,1)}, anchor=north west, legend columns=3, font=\small},
        ]

        \addplot[
            scatter, only marks, mark=*, mark size=2pt, color=blue,
            xshift=-3mm, 
            error bars/.cd, y dir=both, y explicit, error bar style={color=blue}, error mark=-,
        ] coordinates {
            (0, 2.47) += (0, 0.24) -= (0, 0.21)
        };
        \addplot[
            scatter, only marks, mark=square*, mark size=2pt, color=orange,
            xshift=0mm, 
            error bars/.cd, y dir=both, y explicit, error bar style={color=orange}, error mark=-,
        ] coordinates {
            (0, 2.32) += (0, 0.76) -= (0, 0.56)
        };
        \addplot[
            scatter, only marks, mark=triangle*, mark size=2pt, color=green!60!black,
            xshift=3mm, 
            error bars/.cd, y dir=both, y explicit, error bar style={color=green!60!black}, 
            error mark=-,
        ] coordinates {
            (0, 2.42) += (0, 0.44) -= (0, 0.66)
        };

        \addplot[
            scatter, only marks, mark=*, mark size=2pt, color=blue,
            xshift=-3mm, 
            error bars/.cd, y dir=both, y explicit, error bar style={color=blue}, error mark=-,
        ] coordinates {
            (1, 2.61) += (0, 0.31) -= (0, 0.33)
        };
        \addplot[
            scatter, only marks, mark=square*, mark size=2pt, color=orange,
            xshift=0mm, 
            error bars/.cd, y dir=both, y explicit, error bar style={color=orange},  error mark=-,
        ] coordinates {
            (1, 2.22) += (0, 0.1) -= (0, 0.08)
        };
        \addplot[
            scatter, only marks, mark=triangle*, mark size=2pt, color=green!60!black,
            xshift=3mm, 
            error bars/.cd, y dir=both, y explicit, error bar style={color=green!60!black}, 
            error mark=-,
        ] coordinates {
            (1, 2.32) += (0, 0.21) -= (0, 0.15)
        };

        \addplot[
            scatter, only marks, mark=*, mark size=2pt, color=blue,
            xshift=-3mm, 
            error bars/.cd, y dir=both, y explicit, error bar style={color=blue}, error mark=-,
        ] coordinates {
            (2, 2.72) += (0, 0.07) -= (0, 0.09)
        };
        \addplot[
            scatter, only marks, mark=square*, mark size=2pt, color=orange,
            xshift=0mm, 
            error bars/.cd, y dir=both, y explicit, error bar style={color=orange},  error mark=-,
        ] coordinates {
            (2, 2.76) += (0, 0.42) -= (0, 0.22)
        };
        \addplot[
            scatter, only marks, mark=triangle*, mark size=2pt, color=green!60!black,
            xshift=3mm, 
            error bars/.cd, y dir=both, y explicit, error bar style={color=green!60!black}, 
            error mark=-,
        ] coordinates {
            (2, 2.48) += (0, 0.23) -= (0, 0.24)
        };

        \addplot[
            scatter, only marks, mark=*, mark size=2pt, color=blue,
            xshift=-3mm, 
            error bars/.cd, y dir=both, y explicit, error bar style={color=blue}, error mark=-,
        ] coordinates {
            (3, 3.57) += (0, 0.42) -= (0, 0.22)
        };
        \addplot[
            scatter, only marks, mark=square*, mark size=2pt, color=orange,
            xshift=0mm, 
            error bars/.cd, y dir=both, y explicit, error bar style={color=orange},  error mark=-,
        ] coordinates {
            (3, 3.31) += (0, 0.59) -= (0, 0.62)
        };
        \addplot[
            scatter, only marks, mark=triangle*, mark size=2pt, color=green!60!black,
            xshift=3mm, 
            error bars/.cd, y dir=both, y explicit, error bar style={color=green!60!black}, 
            error mark=-,
        ] coordinates {
            (3, 2.47) += (0, 0.19) -= (0, 0.31)
        };

        \addplot[
            scatter, only marks, mark=*, mark size=2pt, color=blue,
            xshift=-3mm, 
            error bars/.cd, y dir=both, y explicit, error bar style={color=blue}, error mark=-,
        ] coordinates {
            (4, 6.79) += (0, 0.31) -= (0, 0.26)
        };
        \addplot[
            scatter, only marks, mark=square*, mark size=2pt, color=orange,
            xshift=0mm, 
            error bars/.cd, y dir=both, y explicit, error bar style={color=orange},  error mark=-,
        ] coordinates {
            (4, 5.01) += (0, 0.28) -= (0, 0.19)
        };
        \addplot[
            scatter, only marks, mark=triangle*, mark size=2pt, color=green!60!black,
            xshift=3mm, 
            error bars/.cd, y dir=both, y explicit, error bar style={color=green!60!black}, 
            error mark=-,
        ] coordinates {
            (4, 2.96) += (0, 0.14) -= (0, 0.2)
        };
    
        \addlegendentry{$\beta$ = 0.5}
        \addlegendentry{$\beta$ = 0.7}
        \addlegendentry{$\beta$ = 0.9}
        
        \end{axis}
        
    \end{tikzpicture}

    \caption{Estimated energy consumption in SNN-HDC models.}
    \label{fig:snn-hdc-dimensionality-beta-synaptic-operations}
  \end{subfigure}
  \caption{SNN-HDC results trained on the DvsGesture dataset \cite{amir_low_2017} for various dimensionality and membrane potential decay rate ($\beta$) values.}
  \label{fig:snn-hdc-accuracy-energy-combined}
\end{figure}

Figure \ref{fig:snn-hdc-dimensionality-beta-accuracy} shows the accuracy of SNN-HDC models trained using different dimensionality and $\beta$ values. In general increasing the number of dimensions leads to a higher final accuracy. The highest accuracy $11$ dimensions is $84.85\%$ ($\beta=0.5$) and the highest accuracy for $2048$ dimensions is $96.59\%$ ($\beta=0.9$). This is expected as higher dimension hypervectors are more robust and can contain a higher number of errors before losing their meaning. The value of $\beta$ does not affect accuracy significantly but it does affect the consistency of the accuracy achieved beyond $32$ dimensions, where $\beta=0.9$ offers better consistency than other values. The trends show that all three $\beta$ values tested have reached a plateau with increasing dimensionality values.

Figure \ref{fig:snn-hdc-dimensionality-beta-synaptic-operations} shows the estimated energy used by various SNN-HDC models trained using different dimensionality and $\beta$ values. Details of how energy consumption is estimated are seen in section \ref{sub:estimated-energy}. Increasing the number of dimensions results in higher estimated energy consumption. This is expected as higher dimensionality values require more neurons and synapses. The most notable trend here is the way the estimated energy consumption increases for increased dimensionality values. For $\beta=0.5$ and $\beta=0.7$ the estimated energy consumption increases exponentially. When $\beta=0.5$, $11$ dimensions consume an average of $2.47$ mJ and $2048$ dimensions consume an average of $6.79$ mJ. For $\beta=0.9$ the estimated energy consumption increases linearly, although dimensionality values beyond $2048$ may show that the trend is still exponential albeit at a slower rate. When $\beta=0.9$, $11$ dimensions consume an average of $2.42$ mJ and $2048$ dimensions consume an average of $2.96$ mJ. The general trend of higher $\beta$ values using less estimated energy than lower $\beta$ values is expected as a lower $\beta$ value causes the membrane voltage to decay quicker meaning the neuron can only integrate information from a shorter time period. As input spikes need to be temporally close to each other to cause a neuron to spike, during training the model compensates for this by making the input neurons spike more frequently thus resulting in higher energy consumption.

While the plots only show a few dimensions, more dimensions tested. The dimensions that were plotted were just to highlight the trend. It was found during testing that the accuracy of the SNN-HDC model did not improve beyond $1024$ dimensions. While the accuracy did not increase with higher dimensions, the estimated energy consumption did. The SNN-HDC model that will be used for further analysis in this paper has a dimensionality of $1024$ and a $\beta$ value of $0.9$. The additional hidden layer in the deeper versions of the rate and latency decoded SNNs will include $1024$ neurons to be equal to the dimensionality value of the chosen SNN-HDC model.

\subsection{Rate and Latency SNN Analysis}\label{sub:rate_latency_analysis}

\begin{figure}[H]
  \centering
  \begin{subfigure}{\textwidth}
    \centering
    \begin{tikzpicture}
        \begin{axis}[
            width  = 0.6\textwidth,
            ymax=100, 
            ylabel={Accuracy (\%)}, 
            xtick={0,1,2,3}, 
            xticklabels={\makecell{Rate-1 \\ (Same Depth)}, \makecell{Rate-2 \\ (Deeper)}, \makecell{Latency-1 \\ (Same Depth)}, \makecell{Latency-2 \\ (Deeper)}}, 
            xticklabel style={
                font=\small, 
            },
            enlargelimits=0.25, 
            ymajorgrids, 
            grid style=dashed, 
            nodes near coords, 
            nodes near coords style={font=\tiny, black, anchor=south, text opacity=0}, 
            legend style={at={(0,1)}, anchor=north west, legend columns=3, font=\small}, 
        ]

        \addplot[
            scatter, only marks, mark=*, mark size=2pt, color=blue,
            xshift=-3mm, 
            error bars/.cd, y dir=both, y explicit, error bar style={color=blue}, error mark=-,
        ] coordinates {
            (0, 93.94) += (0, 0.38) -= (0, 0.38)
        };
        \addplot[
            scatter, only marks, mark=square*, mark size=2pt, color=orange,
            xshift=0mm, 
            error bars/.cd, y dir=both, y explicit, error bar style={color=orange}, error mark=-,
        ] coordinates {
            (0, 93.69) += (0, 1.01) -= (0, 0.88)
        };
        \addplot[
            scatter, only marks, mark=triangle*, mark size=2pt, color=green!60!black,
            xshift=3mm, 
            error bars/.cd, y dir=both, y explicit, error bar style={color=green!60!black}, 
            error mark=-,
        ] coordinates {
            (0, 94.19) += (0, 0.88) -= (0, 1.01)
        };

        \addplot[
            scatter, only marks, mark=*, mark size=2pt, color=blue,
            xshift=-3mm, 
            error bars/.cd, y dir=both, y explicit, error bar style={color=blue}, error mark=-,
        ] coordinates {
            (1, 95.45) += (0, 0.0) -= (0, 0.0)
        };
        \addplot[
            scatter, only marks, mark=square*, mark size=2pt, color=orange,
            xshift=0mm, 
            error bars/.cd, y dir=both, y explicit, error bar style={color=orange},  error mark=-,
        ] coordinates {
            (1, 95.96) += (0, 0.25) -= (0, 0.13)
        };
        \addplot[
            scatter, only marks, mark=triangle*, mark size=2pt, color=green!60!black,
            xshift=3mm, 
            error bars/.cd, y dir=both, y explicit, error bar style={color=green!60!black}, 
            error mark=-,
        ] coordinates {
            (1, 96.34) += (0, 1.01) -= (0, 0.51)
        };

        \addplot[
            scatter, only marks, mark=*, mark size=2pt, color=blue,
            xshift=-3mm, 
            error bars/.cd, y dir=both, y explicit, error bar style={color=blue}, error mark=-,
        ] coordinates {
            (2, 71.72) += (0, 1.77) -= (0, 2.4)
        };
        \addplot[
            scatter, only marks, mark=square*, mark size=2pt, color=orange,
            xshift=0mm, 
            error bars/.cd, y dir=both, y explicit, error bar style={color=orange},  error mark=-,
        ] coordinates {
            (2, 75.38) += (0, 3.41) -= (0, 2.27)
        };
        \addplot[
            scatter, only marks, mark=triangle*, mark size=2pt, color=green!60!black,
            xshift=3mm, 
            error bars/.cd, y dir=both, y explicit, error bar style={color=green!60!black}, 
            error mark=-,
        ] coordinates {
            (2, 75.25) += (0, 3.91) -= (0, 5.93)
        };

        \addplot[
            scatter, only marks, mark=*, mark size=2pt, color=blue,
            xshift=-3mm, 
            error bars/.cd, y dir=both, y explicit, error bar style={color=blue}, error mark=-,
        ] coordinates {
            (3, 80.3) += (0, 3.41) -= (0, 1.89)
        };
        \addplot[
            scatter, only marks, mark=square*, mark size=2pt, color=orange,
            xshift=0mm, 
            error bars/.cd, y dir=both, y explicit, error bar style={color=orange},  error mark=-,
        ] coordinates {
            (3, 81.94) += (0, 1.01) -= (0, 1.26)
        };
        \addplot[
            scatter, only marks, mark=triangle*, mark size=2pt, color=green!60!black,
            xshift=3mm, 
            error bars/.cd, y dir=both, y explicit, error bar style={color=green!60!black}, 
            error mark=-,
        ] coordinates {
            (3, 83.21) += (0, 2.02) -= (0, 1.01)
        };
    
        \addlegendentry{$\beta$ = 0.5}
        \addlegendentry{$\beta$ = 0.7}
        \addlegendentry{$\beta$ = 0.9}
        
        \end{axis}
        
    \end{tikzpicture}

    \caption{Accuracy obtained by comparable rate and latency decoded models.}
    \label{fig:rate-latency-beta-accuracy}
  \end{subfigure}
  \hfill
  \begin{subfigure}{\textwidth}
        \centering
    \begin{tikzpicture}
        \begin{axis}[
            width  = 0.6\textwidth,
            ymin=0, 
            ylabel={\makecell{Estimated Energy Consumption \\ Per Sample (mJ)}},
            scaled y ticks = false,
            xtick={0,1,2,3}, 
            xticklabels={\makecell{Rate-1 \\ (Same \\ Depth)}, \makecell{Rate-2 \\ (Deeper)}, \makecell{Latency-1 \\ (Same \\ Depth)}, \makecell{Latency-2 \\ (Deeper)}}, 
            xticklabel style={
                font=\small, 
            },
            enlargelimits=0.25, 
            ymajorgrids, 
            grid style=dashed, 
            nodes near coords, 
            nodes near coords style={font=\tiny, black, anchor=south, text opacity=0}, 
            legend style={at={(0,1)}, anchor=north west, legend columns=3, font=\small}, 
        ]

        \addplot[
            scatter, only marks, mark=*, mark size=2pt, color=blue,
            xshift=-3mm, 
            error bars/.cd, y dir=both, y explicit, error bar style={color=blue}, error mark=-,
        ] coordinates {
            (0, 3.55) += (0, 0.07) -= (0, 0.06)
        };
        \addplot[
            scatter, only marks, mark=square*, mark size=2pt, color=orange,
            xshift=0mm, 
            error bars/.cd, y dir=both, y explicit, error bar style={color=orange}, error mark=-,
        ] coordinates {
            (0, 3.75) += (0, 0.17) -= (0, 0.32)
        };
        \addplot[
            scatter, only marks, mark=triangle*, mark size=2pt, color=green!60!black,
            xshift=3mm, 
            error bars/.cd, y dir=both, y explicit, error bar style={color=green!60!black}, 
            error mark=-,
        ] coordinates {
            (0, 4.35) += (0, 0.05) -= (0, 0.08)
        };

        \addplot[
            scatter, only marks, mark=*, mark size=2pt, color=blue,
            xshift=-3mm, 
            error bars/.cd, y dir=both, y explicit, error bar style={color=blue}, error mark=-,
        ] coordinates {
            (1, 9.96) += (0, 0.84) -= (0, 0.71)
        };
        \addplot[
            scatter, only marks, mark=square*, mark size=2pt, color=orange,
            xshift=0mm, 
            error bars/.cd, y dir=both, y explicit, error bar style={color=orange},  error mark=-,
        ] coordinates {
            (1, 9.68) += (0, 0.53) -= (0, 0.7)
        };
        \addplot[
            scatter, only marks, mark=triangle*, mark size=2pt, color=green!60!black,
            xshift=3mm, 
            error bars/.cd, y dir=both, y explicit, error bar style={color=green!60!black}, 
            error mark=-,
        ] coordinates {
            (1, 9.67) += (0, 0.34) -= (0, 0.36)
        };

        \addplot[
            scatter, only marks, mark=*, mark size=2pt, color=blue,
            xshift=-3mm, 
            error bars/.cd, y dir=both, y explicit, error bar style={color=blue}, error mark=-,
        ] coordinates {
            (2, 3.11) += (0, 0.15) -= (0, 0.1)
        };
        \addplot[
            scatter, only marks, mark=square*, mark size=2pt, color=orange,
            xshift=0mm, 
            error bars/.cd, y dir=both, y explicit, error bar style={color=orange},  error mark=-,
        ] coordinates {
            (2, 2.78) += (0, 0.19) -= (0, 0.24)
        };
        \addplot[
            scatter, only marks, mark=triangle*, mark size=2pt, color=green!60!black,
            xshift=3mm, 
            error bars/.cd, y dir=both, y explicit, error bar style={color=green!60!black}, 
            error mark=-,
        ] coordinates {
            (2, 3.35) += (0, 0.35) -= (0, 0.2)
        };

        \addplot[
            scatter, only marks, mark=*, mark size=2pt, color=blue,
            xshift=-3mm, 
            error bars/.cd, y dir=both, y explicit, error bar style={color=blue}, error mark=-,
        ] coordinates {
            (3, 7.76) += (0, 0.56) -= (0, 0.79)
        };
        \addplot[
            scatter, only marks, mark=square*, mark size=2pt, color=orange,
            xshift=0mm, 
            error bars/.cd, y dir=both, y explicit, error bar style={color=orange},  error mark=-,
        ] coordinates {
            (3, 7.74) += (0, 0.43) -= (0, 0.26)
        };
        \addplot[
            scatter, only marks, mark=triangle*, mark size=2pt, color=green!60!black,
            xshift=3mm, 
            error bars/.cd, y dir=both, y explicit, error bar style={color=green!60!black}, 
            error mark=-,
        ] coordinates {
            (3, 6.88) += (0, 0.48) -= (0, 0.35)
        };
    
        \addlegendentry{$\beta$ = 0.5}
        \addlegendentry{$\beta$ = 0.7}
        \addlegendentry{$\beta$ = 0.9}
        
        \end{axis}
        
    \end{tikzpicture}

    \caption{Estimated energy consumption in comparable rate and latency decoded models.}
    \label{fig:rate-latency-beta-synaptic-operations}
  \end{subfigure}
  \caption{Rate and latency decoded results trained on the DvsGesture dataset \cite{amir_low_2017}  for various membrane potential decay rates ($\beta$).}
  \label{fig:rate-latency-accuracy-energy-combined}
\end{figure}
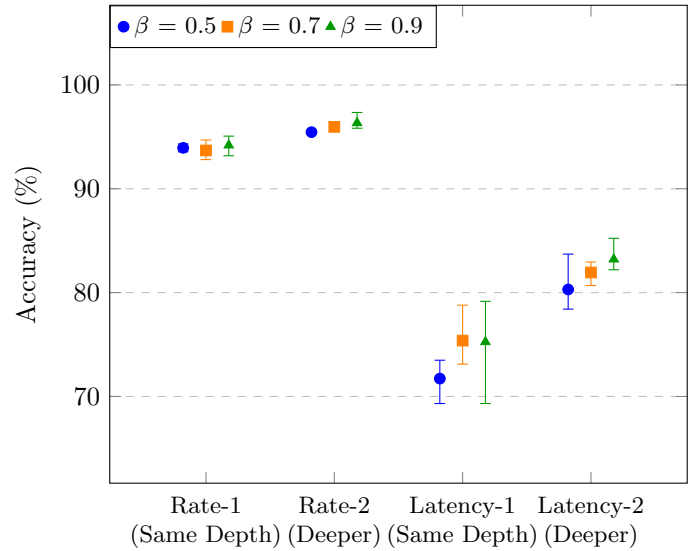
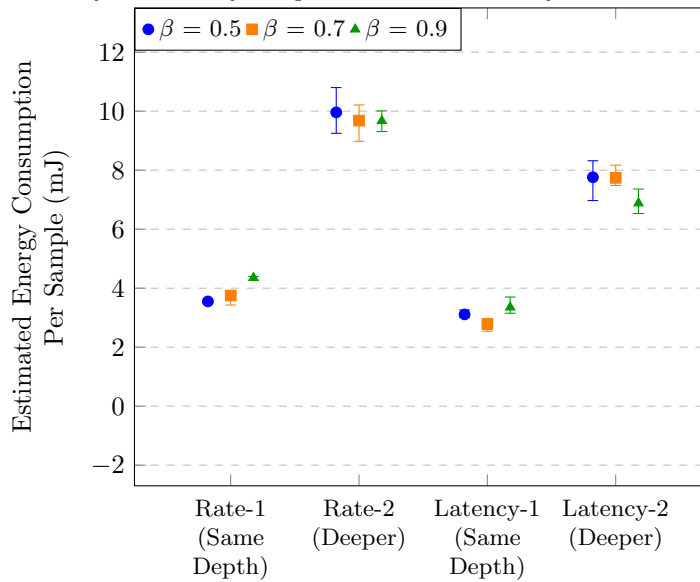

Figure \ref{fig:rate-latency-beta-accuracy} shows the accuracy achieved by the Rate-1, Rate-2, Latency-1 and Latency-2 models on the DvsGesture dataset \cite{amir_low_2017}. The rate decoded models attain higher accuracy than the latency decoded models. The most accurate rate decoded SNN (Rate-2, $\beta=0.9$) achieved $97.35\%$ while the most accurate latency decoded SNN (Latency-2, $\beta=0.9$) achieved $85.23\%$. This is expected as rate decoding typically outperforms latency decoding. The deeper versions of both decoding methods appear to consistently attain higher accuracy than their shallower counterparts. This is expected as deeper networks with more parameters typically attain higher accuracy. The value of $\beta$ has negligible effect on the accuracy of the rate decoded models. The $\beta$ impacts the accuracy of the latency decoded models, with higher values attaining higher accuracy. A possible explanation for this is that rate decoding, unlike latency decoding, is not dependent on the timing of individual spikes but rather the accumulation of many spikes. The latency decoded models have higher ranges of accuracy compared to the rate decoded models likely due to latency decoding being unstable due to its reliance on a single spike for classification.

Figure \ref{fig:rate-latency-beta-synaptic-operations} shows the estimated energy consumption of the Rate-1, Rate-2, Latency-1 and Latency-2 models on the DvsGesture dataset \cite{amir_low_2017}. Details of how energy consumption is estimated are seen in section \ref{sub:estimated-energy}. The rate decoded models used more estimated energy than the latency decoded models which is expected as the higher energy consumption of rate decoding compared to latency decoding is typically seen as one of its drawbacks. When $\beta=0.9$, Rate-2 consumes an average of $9.67$ mJ, while Latency-2 consumes an average of $6.88$ mJ. The deeper version of each model used more estimated energy than its shallower counterpart which is expected as the deeper models contain more neurons and synapses. The value of $\beta$ has negligible impact on estimated energy consumption in all models.

\subsection{Model Activity}\label{sub:model_activity}
The highest accuracy model from each of the five categories of models are used here for further analysis. For the SNN-HDC model this is a dimension count of $1024$ and a $\beta$ value of $0.9$. For the Rate-1 SNN this is a $\beta$ value of $0.9$. For the Rate-2 SNN this is a hidden layer size of $1024$ neurons and a $\beta$ value of $0.9$. For the Latency-1 SNN this is a $\beta$ value of $0.9$. For the Latency-2 SNN this is a hidden layer size of $1024$ and a $\beta$ value of $0.9$. A comparison of relevant metrics for all of these models can be seen in Table \ref{tab:gestures_all_metrics}.

\begin{figure}[H]
  \centering
  \begin{subfigure}{0.48\textwidth}
    \centering
    \begin{tikzpicture}
        \begin{axis}[
            width  = \textwidth,
            height = 8cm,
            major x tick style = transparent,
            ybar=2*\pgflinewidth,
            bar width=6pt,
            ymajorgrids = true,
            ylabel = {Average Per Layer Spike Firing},
            symbolic x coords={Layer 1,Layer 2,Layer 3,Layer 4},
            xtick = data,
            scaled y ticks = false,
            enlarge x limits=0.25,
            ymin=0,
            legend cell align=left,
            legend style={
                    at={(1,1)},
                    anchor=north east,
                    column sep=1ex
            }
            ]
            \addplot[style={green,fill=green,mark=none}]
                coordinates {(Layer 1, 125199.60227273) (Layer 2, 13773.97348485) (Layer 3, 13792.91287879) (Layer 4, 0)};

            \addplot[pattern=crosshatch, draw=blue]
                 coordinates {(Layer 1, 274449.78787879) (Layer 2, 184444.11742424) (Layer 3, 4334.17424242) (Layer 4, 0) };

            \addplot[pattern=north east lines, draw=cyan]
                 coordinates {(Layer 1, 320818.07954545) (Layer 2, 151290.88636364) (Layer 3, 36358.09090909) (Layer 4, 4288.99242424)};

            \addplot[pattern=crosshatch dots, draw=orange]
                 coordinates {(Layer 1, 196352.88257576) (Layer 2, 69541.16666667) (Layer 3, 66.08712121) (Layer 4, 0)};

            \addplot[pattern=grid, draw=magenta]
                 coordinates {(Layer 1, 204905.51893939) (Layer 2, 146252.17424242) (Layer 3, 41612.73106061) (Layer 4, 88.39015152)};

            \legend{SNN-HDC, Rate-1, Rate-2, Latency-1, Latency-2}
        \end{axis}
    \end{tikzpicture}
    \caption{Average per layer spikes fired in the highest accuracy models.}
    \label{fig:gestures_layer_spikes}
  \end{subfigure}
  \hfill
  \begin{subfigure}{0.48\textwidth}
    \centering
    \begin{tikzpicture}
        \begin{axis}[
            width  = \textwidth,
            height = 8cm,
            major x tick style = transparent,
            ybar=2*\pgflinewidth,
            bar width=6pt,
            ymajorgrids = true,
            ylabel = {Per Layer Estimated Energy Consumption (mJ)},
            symbolic x coords={Layer 1,Layer 2,Layer 3,Layer 4},
            xtick = data,
            scaled y ticks = false,
            enlarge x limits=0.25,
            ymin=0,
            legend cell align=left,
            legend style={
                    at={(1.2, 1)},
                    anchor=north east,
                    column sep=1ex,
            }
            ]
            
            \addplot[style={green,fill=green,mark=none}]
                coordinates {(Layer 1, 0.55) (Layer 2, 1.62) (Layer 3, 0.37) (Layer 4, 0)};
    
            \addplot[pattern=crosshatch, draw=blue]
                coordinates {(Layer 1, 0.55) (Layer 2, 3.79) (Layer 3, 0.05) (Layer 4, 0)};

            \addplot[pattern=north east lines, draw=cyan]
                coordinates {(Layer 1, 0.55) (Layer 2, 4.68) (Layer 3, 4.03) (Layer 4, 0.04)};

            \addplot[pattern=crosshatch dots, draw=orange]
                coordinates {(Layer 1, 0.55) (Layer 2, 2.57) (Layer 3, 0.02) (Layer 4, 0)};

            \addplot[pattern=grid, draw=magenta]
                coordinates {(Layer 1, 0.55) (Layer 2, 2.9) (Layer 3, 3.89) (Layer 4, 0.01)};
    
            \legend{SNN-HDC, Rate-1, Rate-2, Latency-1, Latency-2}
        \end{axis}
    \end{tikzpicture}
    \caption{Per layer estimated energy consumption in the highest accuracy models.}
    \label{fig:gestures_layer_synaptic_operations}
  \end{subfigure}
  \caption{SNN-HDC, rate decoded and latency decoded layer activity. Numbers obtained over DvsGesture \cite{amir_low_2017} test set.}
  \label{fig:snn-hdc-rate-latency-per-layer-activity}
\end{figure}
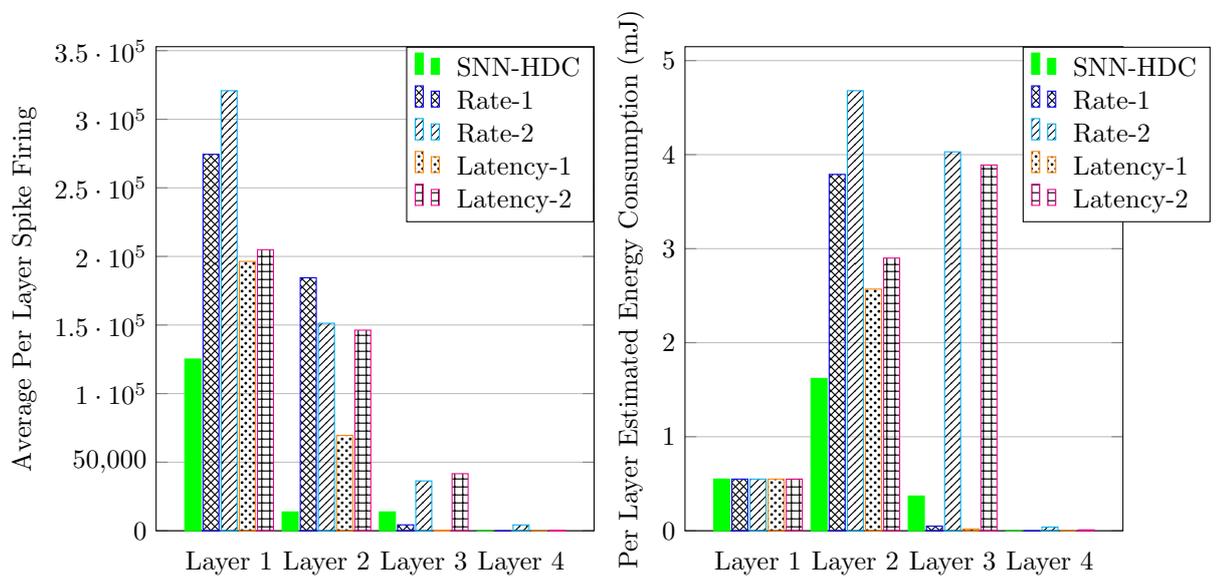

Figure \ref{fig:gestures_layer_spikes} shows the per sample average number of spikes fired by each layer for all five models when passed the entire DvsGesture \cite{amir_low_2017} test set. The first two layers of each network are structurally identical, however, the SNN-HDC model fired significantly fewer spikes than all other models. The rate decoded models fire more spikes than the SNN-HDC as the rate decoded loss function necessitates significantly more output spikes than the hyperdimensional loss function. The latency decoded models fire more spikes than the SNN-HDC, possibly because a large number of spikes over a relatively short time period within the network are needed to help differentiate between classes when using a loss function based around the earliest possible firing of a single output spike.

Figure \ref{fig:gestures_layer_synaptic_operations} shows the estimated energy consumption in each layer of every model. The energy consumption is relative to the number of spikes processed and the architecture of the network. In Layer 1, the models all use the exact same amount of energy as the first layer in all models has the same structure and takes the exact same inputs. In Layer 2 the SNN-HDC model used the least energy which is expected as this layer has a common structure between all models and the SNN-HDC had to accumulate the least number of spikes. In Layer 3 the SNN-HDC consumed more energy than Rate-1 and Latency-1 which is expected as the size of the layers is significantly different. In Layer 3 the SNN-HDC has $1024$ neurons while Rate-1 and Latency-1 only have $11$ neurons, resulting in more accumulations and thereby more energy consumption in that layer of the SNN-HDC. While the SNN-HDC has over $93\times$ more neurons in Layer 3, compared to Rate-1 and Latency-1, the energy consumption only increased by $7.4\times$ and $18.5\times$ respectively.

\begin{figure}[H]
    \centering
    \begin{tikzpicture}
        \begin{axis}[
            width  = 0.48*\textwidth,
            height = 8cm,
            major x tick style = transparent,
            ybar=2*\pgflinewidth,
            bar width=6pt,
            ymajorgrids = true,
            ylabel = {Average Per Layer Firing Rates (Hz)},
            symbolic x coords={Layer 1,Layer 2,Layer 3,Layer 4},
            xtick = data,
            scaled y ticks = false,
            enlarge x limits=0.25,
            ymin=0,
            legend cell align=left,
            legend style={
                    at={(0, 1)},
                    anchor=north west,
                    column sep=1ex
            }
            ]
            
            \addplot[style={green,fill=green,mark=none}]
                coordinates {(Layer 1, 26.6) (Layer 2, 11.5) (Layer 3, 9.0) (Layer 4, 0)};
    
            \addplot[pattern=crosshatch, draw=blue]
                coordinates {(Layer 1, 58.3) (Layer 2, 153.7) (Layer 3, 262.7) (Layer 4, 0)};

            \addplot[pattern=north east lines, draw=cyan]
                coordinates {(Layer 1, 68.2) (Layer 2, 126.1) (Layer 3, 23.7) (Layer 4, 260.0)};

            \addplot[pattern=crosshatch dots, draw=orange]
                coordinates {(Layer 1, 41.7) (Layer 2, 58.0) (Layer 3, 4.0) (Layer 4, 0)};

            \addplot[pattern=grid, draw=magenta]
                coordinates {(Layer 1, 43.6) (Layer 2, 121.9) (Layer 3, 27.1) (Layer 4, 5.4)};
    
            \legend{SNN-HDC, Rate-1, Rate-2, Latency-1, Latency-2}
        \end{axis}
    \end{tikzpicture}
    \caption{Average per layer firing rate of neurons in each model per sample of the DvsGesture \cite{amir_low_2017} test set.}
    \label{fig:average_firing_rate_of_models}
\end{figure}
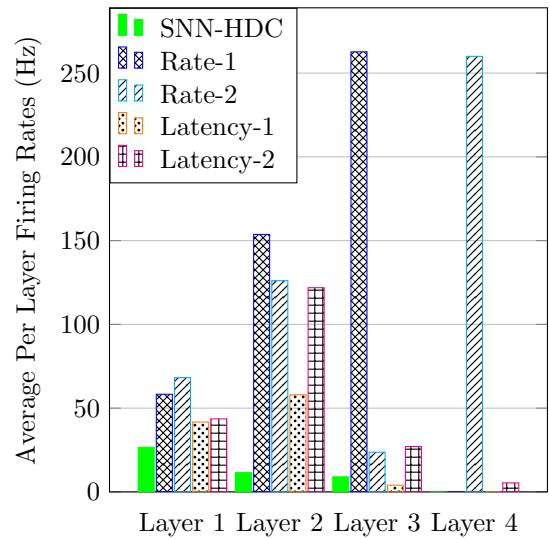

Figure \ref{fig:average_firing_rate_of_models} shows the average per sample firing rate (Hz) of all models per layer. Table \ref{tab:average_firing_rate_of_models} show the average per sample firing rate (Hz) of each model as a whole. It is clear that the rate models have the highest firing rate, followed by the latency models. The SNN-HDC has the lowest firing rate of all models, being $2.19 \times$ lower than the second best (Latency-1).

\begin{table}[H]
    \centering
    \caption{Average firing rate of neurons in each model per sample of the DvsGesture \cite{amir_low_2017} test set.}
    \begin{tabular}{||c | c | c | c||}
        \hline
        Model & \makecell{Number of \\ Neurons} & \makecell{Average Firing \\ Rate (Hz)} & \makecell{Relative \\ Firing Rate} \\
        \hline \hline
        SNN-HDC & 4960 & 20.5 & $1\times$ \\
        \hline
        Rate-1 & 3947 & 78.2 & $3.81\times$ \\
        \hline
        Rate-2 & 4971 & 68.2 & $3.33\times$ \\
        \hline
        Latency-1 & 3947 & 44.9 & $2.19\times$ \\
        \hline
        Latency-2 & 4971 & 52.7 & $2.57\times$ \\
        \hline
    \end{tabular}

    \label{tab:average_firing_rate_of_models}
\end{table}

\subsection{Model Latency}\label{sub:model_latency}

\begin{table}[H]
    \centering
    \caption{Latency of models on DvsGesture \cite{amir_low_2017} test set shown as average $\pm$ standard deviation.}
    \begin{tabular}{||c|c|c||}
        \hline
        Model & Latency & Relative\\
        \hline\hline
        SNN-HDC & $162 \, \mathrm{ms} \pm 278 \, \mathrm{ms}$ & $1\times$ \\
        \hline
        Rate-1 & $208 \, \mathrm{ms} \pm 299 \, \mathrm{ms}$ & $1.28\times$ \\
        \hline
        Rate-2 & $133 \, \mathrm{ms} \pm 218 \, \mathrm{ms}$ & $0.82\times$ \\
        \hline
        Latency-1 & $206 \, \mathrm{ms} \pm 311 \, \mathrm{ms}$ & $1.27\times$ \\
        \hline
        Latency-2 & $188 \, \mathrm{ms} \pm 234 \, \mathrm{ms}$ & $1.16\times$ \\
        \hline
    \end{tabular}
    \label{tab:gestures-latencies}
\end{table}

Table \ref{tab:gestures-latencies} shows the latency achieved by each of the five models on the DvsGesture \cite{amir_low_2017} test set. The latency is shown as an average $\pm$ standard deviation in milliseconds. Latency decoding is typically touted as having improved latency compared to rate decoding, however, the results here show this is not the case. Whilst this is counter intuitive given how the loss functions are setup, it is not unexpected as prior research has already shown that rate decoding can attain better latency than latency decoding when trained on neuromorphic datasets \cite{liu_first-spike_2023}. The deeper models for both rate and latency decoding exhibit lower latency than their shallower counterparts. A possible explanation for this is that the additional hidden layer helped to increase class separability, as evidenced by both deeper models also having better accuracy than their shallower counterparts. The latency of the SNN-HDC is better and more consistent than Rate-1, but it is slower and less consistent than Rate-2. This again is likely due to structural differences, with deeper networks offering improved performance.

\subsection{Unknown Class Detection}\label{sub:unknown_class}
The eleventh class in the DvsGesture dataset \cite{amir_low_2017} is the random class. This class was removed from both the training and testing datasets. The SNN-HDC model trained here has the same parameters as the chosen model from the full dataset. The dimensionality count is $1024$ and the $\beta$ value is $0.9$. After three different runs (affecting weight initialisation and generated class hypervectors), the average accuracy was $96.67\%$ with a range of $\pm0.83\%$ when classifying only the ten classes that were used for training. The highest accuracy of these models was $97.08\%$, which is used for further analysis in this section. 

\begin{figure}[H]
    \centering
    \begin{tikzpicture}
        \begin{axis}[
            width = 0.6\textwidth,
            xlabel={$\delta$ threshold},
            ylabel={Accuracy (\%)},
            xmin=0.1, xmax=0.4,
            ymin=0, ymax=100,
            xtick={0.1, 0.15, 0.2, 0.25, 0.3, 0.35, 0.4},
            grid=both,
            legend style={at={(0.0,0.0)}, anchor=south west},
            major grid style={dashed,gray!50},
            minor grid style={dotted,gray!20},
            xticklabel style={font=\small},
            yticklabel style={font=\small},
            axis on top,
            axis lines=left,
            ]

            \addplot[
                color=blue,
                line width=1pt,
            ] coordinates {
                (0.4, 87.88)
                (0.35, 88.64)
                (0.3, 90.15)
                (0.25, 91.29)
                (0.2, 89.39)
                (0.15, 85.23)
                (0.1, 78.03)
            };
            \addlegendentry{Full Dataset}

            \addplot[
                color=red,
                line width=1pt,
                dashed
            ] coordinates {
                (0.4, 96.67)
                (0.35, 96.67)
                (0.3, 96.25)
                (0.25, 94.17)
                (0.2, 90.83)
                (0.15, 84.17)
                (0.1, 75.83)
            };
            \addlegendentry{Ten Classes}

            \addplot[
                color=green!60!black,
                line width=1pt,
                dotted
            ] coordinates {
                (0.4, 0)
                (0.35, 8.33)
                (0.3, 29.17)
                (0.25, 62.5)
                (0.2, 75)
                (0.15, 95.83)
                (0.1, 100)
            };
            \addlegendentry{Unknown Class}

        \end{axis}
    \end{tikzpicture}
    \caption{Accuracy obtained by the 10-class trained SNN-HDC when performing classification on the full 11 class DvsGesture dataset \cite{amir_low_2017}. A known classification is correct if its normalised Hamming distance is less than $\delta$ to the correct class hypervector. An unknown classification is correct if its normalised Hamming distance is more than $\delta$ to all known class hypervectors.}
    \label{fig:unknown_class_accuracy}
\end{figure}
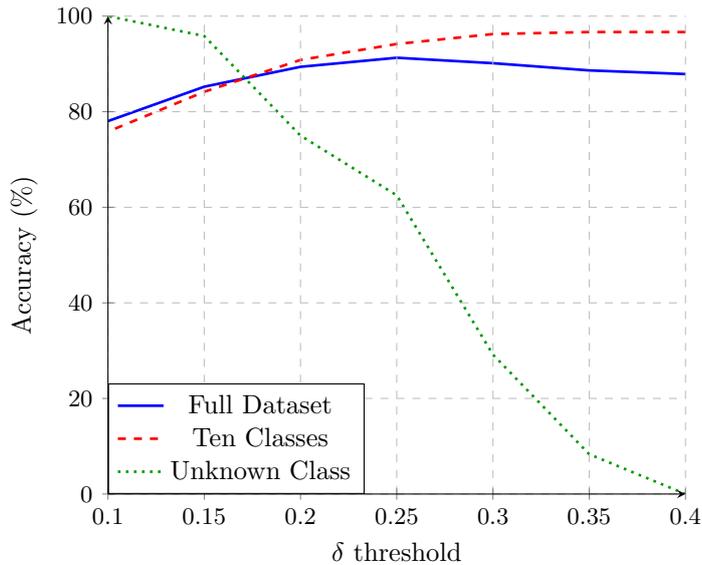

Figure \ref{fig:unknown_class_accuracy} shows the accuracy obtained by the SNN-HDC model trained on only 10 classes, but used to classify all 11 classes. The $\delta$ values refer to normalised Hamming distance (Equation \ref{eq:normalised_hamming_distance}). Classifications of known classes are only counted as true if they are both correct and have a normalized Hamming distance that is less than $\delta$. Classifications of the unknown class are only counted as true when the output has a normalised Hamming distance of more than $\delta$ to every known class. The full dataset accuracy shows the accuracy obtained by the model given both of these constraints. The unknown class accuracy only shows the accuracy of identifying the unknown class. As the $\delta$ value lowers, the models ability to identify the unknown class improves. When $\delta=0.4$ the model is not able to identify any unknown samples, however, when $\delta=0.1$ the model is able to identify every unknown sample. Lowering the value of $\delta$ below $0.25$ appears to have a negative impact on the full dataset accuracy. This is because the requirements of the output are more strict and necessitate fewer errors. The ideal value of $\delta$ in these results is $0.25$ as it is able to identify $62.5\%$ of unknown samples while still achieving a high accuracy of $94.17\%$.

The results here indicate that while it is better to train a model on all 11 classes, training on only 10 classes results in high performance. With a carefully specified $\delta$ value the SNN-HDC model is capable of classifying known and unknown samples to such a high accuracy that it outperforms SQUAT \cite{venkatesh_squat_2024} (the study which inspired the architecture of the models developed here) and both state-of-the-art SNN HDC combinations \cite{zou_memory-inspired_2022, morris_hyperspike_2022} reviewed in section \ref{related-works}.

\subsection{Overall DvsGesture Results}\label{sub:gesture_results}

Table \ref{tab:gestures_all_metrics} shows the performance of all models trained here in terms of accuracy achieved, estimated energy consumption and latency. The SNN-HDC achieves at least a $1.51\%$ higher accuracy than Rate-1, Latency-1 and Latency-2 but achieves $0.76\%$ less than Rate-2. The SNN-HDC has the smallest spike count of all the models, with reductions ranging from $1.74\times$ to $3.36\times$. The SNN-HDC has the smallest energy consumption with reductions ranging from $1.24\times$ to $3.67\times$. The SNN-HDC achieves at least a $1.16\times$ lower latency than Rate-1, Latency-1 and Latency-2 but has $1.22\times$ higher latency than Rate-2. 

\begin{table}[H]
    \centering
    \caption{Comparison between the proposed SNN-HDC, two rate decoded SNNs and two latency decoded SNNs. Accuracy, spike count and estimated energy consumption were obtained after each of the networks trained to convergence on the DvsGesture dataset \cite{amir_low_2017}.}
    \resizebox{\linewidth}{!}{
    \begin{tabular}{||c | c | c | c| c| c| c| c| c||} 
        \hline
        Model & \makecell{Parameter \\ Number} & Accuracy & \makecell{Average \\ Spike \\ Count} & \makecell{Relative \\ Spike \\ Count} & \makecell{Estimated \\ Energy} & \makecell{Relative \\ Estimated \\ Energy} & \makecell{Latency} & \makecell{Relative \\ Latency}\\ [0.5ex]
        \hline\hline
        
        SNN-HDC & $834$ k & $96.59\%$ & \num{1.53e5} & $1\times$ & \num{2.54} mJ & $1\times$ &  $162 \, \mathrm{ms}$ & $1\times$\\
        \hline
        \makecell{Rate-1 \\ (Same Depth)} & $22.5$ k & $95.08\%$ & \num{4.63e5} & $3.03\times$ & \num{4.39} mJ & $1.73\times$ & $208 \, \mathrm{ms}$ & $1.28\times$\\
        \hline
        \makecell{Rate-2 \\ (Deeper)} & $845$ k & $97.35\%$ & \num{5.13e5} & $3.36\times$ & \num{9.31} mJ & $3.67\times$ & $133 \, \mathrm{ms}$ & $0.82\times$\\
        \hline
        \makecell{Latency-1 \\ (Same Depth)} & $22.5$ k & $79.17\%$ & \num{2.66e5} & $1.74\times$ & \num{3.14} mJ & $1.24\times$ & $206 \, \mathrm{ms}$ & $1.27\times$\\
        \hline
        \makecell{Latency-2 \\ (Deeper)} & $845$ k & $85.23\%$ & \num{3.93e5} & $2.57\times$ & \num{7.37} mJ & $2.90\times$ & $188 \, \mathrm{ms}$ & $1.16\times$\\
        \hline
    \end{tabular}
    }

    \label{tab:gestures_all_metrics}
\end{table}

Table \ref{tab:gestures_performance_from_other works} compares the models tested here with other works in the literature. Rate-1 and the SQUAT model \cite{venkatesh_squat_2024} have the same architecture, however Rate-1 is trained using higher temporal resolution. Rate-1 achieves an $8.84\%$ higher accuracy which indicates, as discussed in Section \ref{resolution_of_data}, that the use of a low temporal resolution diminishes temporal information and thus hinders performance. The table also shows that the SNN-HDC model achieves a higher accuracy than the related state-of-the-art models, achieving $8.79\%$ more than SpikeHD \cite{zou_memory-inspired_2022} and $9.39\%$ more than HyperSpike \cite{morris_hyperspike_2022}.

\begin{table}[H]
    \centering
    \caption{Accuracy obtained by the models in this work (top) and other SNN models (bottom) trained on the DvsGesture dataset \cite{amir_low_2017}.}
    \begin{tabular}{||c | c | c | c | c||} 
        \hline
        Model & \makecell{Parameter \\ Number} & \makecell{Model \\ Type} & \makecell{Decoding \\ Method} & Accuracy \\ [0.5ex] 
        \hline\hline

        Rate-2 & $845$ k & SNN & Rate & $97.35\%$ \\
        \hline
        SNN-HDC & $834$ k & SNN & HDC &  $96.59\%$ \\
        \hline
        Rate-1 & $22.5$ k & SNN & Rate & $95.08\%$ \\
        \hline
        Latency-2 & $845$ k & SNN & Latency & $85.23\%$ \\
        \hline
        Latency-1 & $22.5$ k & SNN & Latency & $79.17\%$ \\
        \hline\hline
        
        ACE-BET \cite{liu_fast_2022} & $11.7$ M & ANN & - & $98.88\%$ \\
        \hline
        mMND \cite{she_sequence_2021} & $1.1$ M & SNN & Rate & $98.0\%$ \\
        \hline
        DECOLLE \cite{kaiser_synaptic_2020} & $1.64$ M & SNN & Rate & $95.54\%$ \\
        \hline
        SLAYER \cite{shrestha_slayer_2018} & $1.06$ M & SNN & Rate & $93.64\%$ \\
        \hline
        FS \cite{liu_first-spike_2023} & $740$ k & SNN & Latency & $92.8\%$ \\
        \hline
        SpikeHD \cite{zou_memory-inspired_2022} & $968$ k & SNN & HDC & $87.8\%$ \\
        \hline
        HyperSpike \cite{morris_hyperspike_2022} & - & SNN & HDC & $87.2\%$ \\
        \hline
        SQUAT \cite{venkatesh_squat_2024} & $22.5$ k & SNN & Rate & $86.24\%$ \\
        \hline
        
    \end{tabular}

    \label{tab:gestures_performance_from_other works}
\end{table}

\subsection{SL-Animals Results}\label{sub:animals_results}
Table \ref{tab:animals_all_metrics} shows resulting metrics for all five models trained on the SL-Animals-DVS dataset \cite{vasudevan_sl-animals-dvs_2022}. Table \ref{tab:animals_performance_from_other works} compares the performance of the models in this work to other published works. All networks are trained with $D=1024$ and $\beta=0.9$ as determined by the experiments performed on the DvsGesture dataset \cite{amir_low_2017}. While the trend of rate decoded models achieving higher accuracy than latency decoded models remains, some of the other trends do not. Here the SNN-HDC is the highest accuracy model, which could simply be the result of the robustness of hypervectors. The trend of deeper models achieving higher accuracy than their shallower counterparts does not hold true for the latency decoded networks. This is likely the result of structural differences which appear in the fully connected portion of the networks. Latency-1 has a fully connected structure of [25 - 19]. Latency-2 has a fully connected structure of [25 - 1024 - 19]. The full architectures are shown in Table \ref{tab:animals-architectures}. It is possible that the increased number of neurons feeding into the output layer causes instability for latency decoding. This likely wasn't seen for the DvsGesture dataset \cite{amir_low_2017} training as there was a much smaller structural difference between Latency-1 ([800 - 11]) and Latency-2 ([800 - 1024 - 11]).

\begin{table}[H]
    \small
    \centering
    \caption{Comparison between the proposed SNN-HDC, two rate decoded SNNs and two latency decoded SNNs. Accuracy, spike count and estimated energy consumption were obtained after each of the networks trained to convergence on the SL-Animals-DVS dataset \cite{vasudevan_sl-animals-dvs_2022}.}
    \resizebox{\linewidth}{!}{
    \begin{tabular}{||c | c | c | c| c| c| c| c| c||} 
        \hline
        Model & \makecell{Parameter \\ Number} & Accuracy & \makecell{Average \\ Spike \\ Count} & \makecell{Relative \\ Spike \\ Count} & \makecell{Estimated \\ Energy} & \makecell{Relative \\ Estimated \\ Energy} & \makecell{Latency} & \makecell{Relative \\ Latency}\\ [0.5ex]
        \hline\hline
        
        SNN-HDC & $55.9$ k & $74.13\%$ & \num{6.86e5} & $1\times$ & \num{3.61} mJ & $1\times$ &  $570 \, \mathrm{ms}$ & $1\times$\\
        \hline
        \makecell{Rate-1 \\ (Same Depth)} & $29.8$ k & $70.68\%$ & \num{1.85e6} & $2.70\times$ & \num{8.21} mJ & $2.27\times$ & $668 \, \mathrm{ms}$ & $1.17\times$\\
        \hline
        \makecell{Rate-2 \\ (Deeper)} & $75.4$ k & $72.74\%$ & \num{1.71e6} & $2.49\times$ & \num{7.56} mJ & $2.09\times$ & $592 \, \mathrm{ms}$ & $1.04\times$\\
        \hline
        \makecell{Latency-1 \\ (Same Depth)} & $29.8$ k & $47.34\%$ & \num{9.31e5} & $1.36\times$ & \num{4.97} mJ & $1.38\times$ & $361 \, \mathrm{ms}$ & $0.63\times$\\
        \hline
        \makecell{Latency-2 \\ (Deeper)} & $75.4$ k & $45.12\%$ & \num{1.08e6} & $1.57\times$ & \num{5.43} mJ & $1.50\times$ & $356 \, \mathrm{ms}$ & $0.62\times$\\
        \hline
    \end{tabular}
    }

    \label{tab:animals_all_metrics}
\end{table}

The SNN-HDC exhibits the lowest average spike count with reductions ranging from $1.36\times$ to $2.70\times$. The SNN-HDC also had the lowest energy consumption with reductions ranging from $1.38\times$ to $2.27\times$. These results show the energy savings of the SNN-HDC are agnostic to architecture and dataset.

The latencies of all models here are noticeably higher than the latencies for the DvsGesture dataset \cite{amir_low_2017} seen in Table \ref{tab:gestures-latencies}. This is likely due to the context of the actions being performed with DvsGesture \cite{amir_low_2017} samples taking less time to become distinct than SL-Animals-DVS \cite{vasudevan_sl-animals-dvs_2022} samples. The latency decoded models have the lowest averages, however, this is negated by their lower classification accuracy. The SNN-HDC achieved lower latency than both rate decoded models with reductions of $1.17\times$ and $1.04\times$ over Rate-1 and Rate-2 respectively.

\begin{table}[H]
    \centering
    \caption{Accuracy obtained by the models in this work (top) and other SNN models (bottom) trained on all subsets of the SL-Animals-DVS dataset \cite{vasudevan_sl-animals-dvs_2022}.}
    \begin{tabular}{||c | c | c | c | c||} 
        \hline
        Model & \makecell{Parameter \\ Number} & \makecell{Model \\ Type} & \makecell{Decoding \\ Method} & Accuracy \\ [0.5ex] 
        \hline\hline
        SNN-HDC & $55.9$ k & SNN & HDC & $74.13\% \pm 4.26\%$ \\
        \hline
        Rate-2 & $75.4$ k & SNN & Rate & $72.74\% \pm 3.73\%$ \\
        \hline
        Rate-1 & $29.8$ k & SNN & Rate & $70.68\% \pm 4.00\%$ \\
        \hline
        Latency-1 & $29.8$ k & SNN & Latency & $47.34\% \pm 3.66\%$ \\
        \hline
        Latency-2 & $75.4$ k & SNN & Latency & $45.12\% \pm 4.46\%$ \\
        \hline\hline
        EventRPG \cite{sun_eventrpg_2024} & $11.7$ M & SNN & Rate & $91.59\%$ \\
        \hline
        EvT \cite{sabater_event_2022} & $0.48$ M & ANN & - & $88.12\%$ \\
        \hline
        SL-Animals \cite{vasudevan_sl-animals-dvs_2022} & $1.37$ M & SNN & Rate &  $70.6\% \pm 7.8\%$ \\
        \hline
    \end{tabular}

    \label{tab:animals_performance_from_other works}
\end{table}

\section{Discussion and Conclusion} \label{discusion-conclusion}
This work presents a new SNN decoding method inspired by distributed representations involving the combination of SNNs and HDC. The proposed method is evaluated and compared to rate and latency decoded SNNs trained on neuromorphic datasets. The results show that for comparable models, the SNN-HDC maintains or improves on the accuracy achieved by both other decoding methods and attains generally lower classification latency while always using fewer spikes, indicating the potential for lower energy use. The SNN-HDC achieved estimated energy consumption reductions ranging from $1.24\times$ to $3.67\times$ on the DvsGesture dataset \cite{amir_low_2017} and from $1.38\times$ to $2.27\times$ on the SL-Animals-DVS dataset \cite{vasudevan_sl-animals-dvs_2022}. The SNN-HDC also offers the ability to identify unknown classes that it has not been trained on, being able to identify $100\%$ of the samples from an unseen/untrained class from the DvsGesture dataset \cite{amir_low_2017}.

While the SNN-HDC approach offers multiple benefits over rate and latency decoding, these benefits come with increased memory use when simply replacing the output layer. The SNN-HDC used $37.1\times$ and $1.88\times$ more parameters than the same depth one-hot encoded counterpart on the DvsGesture \cite{amir_low_2017} and SL-Animals-DVS \cite{vasudevan_sl-animals-dvs_2022} datasets respectively. The higher memory footprint can be mitigated by acting on the number of neurons in the layer before the output layer. Should the output layer of a network be removed and its penultimate layer be decoded with hyperdimensional decoding, which is essentially the case between the SNN-HDC model and its deeper one-hot encoded counterparts, the SNN-HDC model will use fewer parameters. The SNN-HDC model used $1.01\times$ and $1.35\times$ fewer parameters than the deeper one-hot encoded counterpart on the DvsGesture \cite{amir_low_2017} and SL-Animals-DVS \cite{vasudevan_sl-animals-dvs_2022} datasets respectively. In the case that the number of classes exceeds $2633$ then the SNN-HDC model will always use fewer parameters than a one-hot encoded counterpart as the expressivity of hypervectors will exceed the expressivity of one-hot encoding as discussed in section \ref{sub:expressiveness}.

The proposed SNN-HDC model advances prior SNN and HDC approaches \cite{zou_memory-inspired_2022, morris_hyperspike_2022} in several aspects. The SNN-HDC is trained directly to produce hypervectors removing the need to initially train the network using other representations. The SNN-HDC also changes the encoding method of hypervectors removing the reliance on matrix multiplication, saving energy by reducing the amount of computation used. The method also lowers the latency at which classifications can be made, as hypervectors are built up over time and can be classified at any time as opposed to only at the end of a sample. Compared to the state-of-the-art \cite{zou_memory-inspired_2022, morris_hyperspike_2022} the SNN-HDC also uses a more energy efficient method for hypervector comparison, in the form of Hamming distance.

The work here has further demonstrated the potential of combining SNNs with HDC and warrants further exploration. Future research on this topic could explore encoding hypervectors to include the number of spikes and inter spike interval into the dimensional values to further increase the information obtained per output. This could help to improve the accuracy of classifications and to lower the energy usage of the network further. The effect of the number of on bits in the class hypervectors could be analysed, as using fewer on bits could potentially lower energy usage. The scalability of the SNN-HDC approach could be explored when applied to problems/data with such a high number of classes that hypervectors are more expressive than one-hot encoding, as discussed in section \ref{sub:expressiveness}. The SNN could also be trained to output dynamic rather than static hypervectors, producing an output that moves over time through hyperspace, enabling an SNN to make continuous classifications over time in an event driven manner without needing to be reset. This opens up applications to continuous online data streams, which would be well suited to neuromorphic deployment.

\funding{This research was supported through the NimbleAI project, funded via the Horizon Europe Research and Innovation programme (Grant Agreement 101070679), and UKRI under the UK government’s Horizon Europe funding guarantee (Grant Agreement 10039070); the Horizon Europe AIDA4Edge project (Grant Agreement 101160293); and the EPSRC Edgy Organism project (EP/Y030133/1).}

\roles{Cedrick Kinavuidi \orcid{0009-0001-2860-9163} 
\par\noindent
Conceptualization (equal), Data curation (lead), Formal analysis (lead), Investigation (lead), Methodology (lead), Project administration (equal), Resources (equal), Software (lead), Validation (lead), Visualization (lead), Writing – original draft (lead), Writing – review \& editing (equal)

\vspace{1em}\par\noindent
Luca Peres \orcid{0000-0001-9748-9073} 
\par\noindent
Conceptualization (equal), Formal analysis (supporting), Funding acquisition (supporting), Investigation (supporting), Methodology (supporting), Project administration (equal), Resources (equal), Software (supporting), Supervision (supporting), Validation (supporting), Visualization (supporting), Writing – original draft (supporting), Writing – review \& editing (equal)

\vspace{1em}\par\noindent
Oliver Rhodes \orcid{0000-0003-1728-2828}
\par\noindent
Conceptualization (equal), Formal analysis (supporting), Funding acquisition (lead), Investigation (supporting), Methodology (supporting), Project administration (equal), Resources (equal), Software (supporting), Supervision (lead), Validation (supporting), Visualization (supporting), Writing – original draft (supporting), Writing – review \& editing (equal)
}

\data{The data that support the findings of this study are openly available at the following URL/DOI: https://doi.org/10.5281/zenodo.17522304.}

\providecommand{\newblock}{}

\end{document}